\documentclass[onecolumn]{robbyant}           %

\metadata[Website]{\url{https://technology.robbyant.com/lingbot-va-v2}}

\newcommand\methodname{LingBot-VA}
\newcommand{\methodold}{\texttt{\methodname}\xspace}
\newcommand{\method}{\texttt{\methodname~2.0}\xspace}

\usepackage{tikz}
\usetikzlibrary{tikzmark, calc}
\usepackage{makecell}

\algblockdefx[MainLoop]{MainLoop}{EndMainLoop}{\textbf{loop}}{\textbf{end loop}}

\title{Native Video-Action Pretraining for \\[4pt] Generalizable Robot Control}

\author{
\begin{center}
    Qihang Zhang,
    Lin Li,
    Luyao Zhang,
    Shuai Yang,
    Yiming Luo,
    Shuaiting Li,
    Ruilin Wang,
    Junke Wang,
    \\[2pt]
    Jiahao Shao,
    Gangwei Xu,
    Jiaming Zhou,
    Yishu Shen,
    Yudong Jin,
    Fangyi Xu,
    Shuailei Ma,
    Jiaqi Liao,
    \\[2pt]
    Guanxing Lu,
    Zifan Shi,
    Yongkun Wen,
    Yujie Zhao,
    Weixuan Tang,
    Xinyang Wang,
    Chaojian Li,
    \\[2pt]
    Jiapeng Zhu,
    Ka Leong Cheng,
    Nan Xue,
    Xing Zhu,
    Yujun Shen,
    Yinghao Xu$^{\dagger}$
    \\[12pt]
    $^{\dagger}$Project Lead
\end{center}
}

\begin{document}

\abstract{%
The advent of video-action models offers a promising path for robot control.
Nevertheless, we argue that repurposing video generative models designed for digital content creation is inherently inadequate for physical environments.
To bridge this gap, we present \method, a video-action foundation model built from the ground up for embodiment.
Four core design principles showcase its evolution from \methodold.
(1) Departing from traditional reconstruction-focused VAEs, we introduce \textbf{\textit{a semantic visual-action tokenizer}}, which aligns visual representations with both semantics and actions, improving instruction following and action precision in subsequent policy learning.
(2) Given the strictly causal nature of temporal dynamics, we adopt \textbf{\textit{a causal pretraining paradigm}}, training from scratch to circumvent the catastrophic forgetting that frequently occurs when adapting bidirectional architectures.
(3) To meet the demands of high-frequency inference, our model employs \textbf{\textit{a sparse MoE backbone}}, expanding model capacity without compromising efficiency.
(4) Real-time closed-loop control is realized through \textbf{\textit{an enhanced asynchronous inference scheme}}, which predicts future latents in parallel with action execution while re-grounding each rollout on the latest observation via learned forward dynamics.
Real-world deployment validates \method as a robust foundation model, as evidenced by its few-shot generalization across complex manipulation tasks.
}

\maketitle

\justifying
\section{Introduction}
\label{sec:introduction}

Video-action models such as LingBot-VA~\cite{VA} and DreamZero~\cite{dreamzero}
have recently emerged as a powerful paradigm for generalist robot manipulation.
Rather than mapping observations directly to actions, as reactive
vision-language-action policies do~\cite{rt2,pi0,kim2024}, they jointly predict
how a scene will evolve and how to act within it, grounding control in physical
dynamics and improving sample efficiency and
generalization~\cite{uwm,uva,vpp}. Much of this capability, however, is inherited from their
video-pretrained backbones, suggesting that generalist robot control depends as
much on the pretrained foundation as on the policy learned upon it.

Current video-action models are still largely built from components designed for
generic video generation---a reconstruction-oriented VAE and a 
bidirectional video-diffusion backbone---with an action module added for
robotics afterward.
This starting point creates three concrete limitations. First, the
representation is optimized for appearance rather than dynamics:
pixel-reconstruction latents preserve visual detail but carry limited semantic
and physical structure, and the separately attached action module leaves world
states and actions in poorly aligned spaces. Second, inference is too slow for
closed-loop control: high-dimensional video tokens and iterative denoising make
first-generation video-action models costly to run at the frequencies real
robots require. Third, the pretraining signal does not scale toward control:
web-scale video is abundant, but generic video objectives do not teach how
actions reshape the world, so the action signal remains tied to expensive robot
data, limiting the control knowledge learned before downstream adaptation.

These limitations are compounded by a structural mismatch: the backbone is
pretrained with bidirectional attention, while closed-loop control unfolds
strictly forward in time. LingBot-VA resolves this mismatch by finetuning the
stack into a causal video-action model, but the retrofit relies on scarce robot
data and may weaken the broad priors acquired from web-scale pretraining. We therefore pursue a
\emph{native} route: pretraining the full stack, including a semantic
visual-action tokenizer and a causal DiT, for robot control on web-scale image
and video data, so
that the broad priors are acquired natively in causal form rather than
inherited and then eroded.

We present \method, which instantiates this native route with a design that
removes the three limitations above. Rather than attaching actions to an off-the-shelf
video generator, \method first builds a shared semantic latent space for
observations and actions, then learns causal dynamics on top of that space. The
first stage is a semantic visual-action tokenizer~\cite{repwam}: beyond
reconstruction, it aligns video latents to a frozen visual foundation model,
producing compact representations that are far more semantic than
pixel-reconstruction latents. It further learns latent actions through
self-supervision, placing world states and actions in the same space so that
even unlabeled web video carries action-relevant supervision. The second stage
is a causal diffusion transformer that predicts future visual latents together
with the latent actions connecting them, conditioned on language instructions
through cross-attention to a frozen pretrained language model. Its causal
next-latent objective provides a self-supervised signal that scales to web
video while matching the temporal structure of closed-loop control, avoiding
the bidirectional-to-causal retrofit that risks eroding pretrained priors; and its
feed-forward layers are instantiated as a sparse mixture-of-experts, preserving
model capacity while reducing the active computation per step for
high-frequency control. Because adjacent chunks are often visually similar
enough that next-chunk supervision alone rewards copying appearance, we
additionally train with multi-chunk prediction (MCP), which supervises several
future chunks at once so that the learned latents encode trajectory-level
dynamics rather than short-term visual continuity.

Even with a semantic latent space and a sparse backbone, however, deployment on
real hardware faces a serial bottleneck: the robot must continue acting while
the world model updates its prediction; otherwise, model latency directly
becomes control latency. We address this with \emph{Foresight Reasoning}, an asynchronous
inference scheme in which future visual latents are predicted in parallel with
action execution. A naive asynchronous scheme can drift by conditioning on stale
imagined latents. \method instead re-grounds each prediction on the latest real
observation through a learned forward-dynamics step of the policy itself.
This shares the concurrent-inference motivation of Harmonic Reasoning in
GEN-0~\cite{gen0}, but grounds the look-ahead in a learned world model and
corrects it with every real observation.

Because its action signal no longer depends on scarce robot demonstrations,
\method acquires control knowledge at the scale of web video rather than at the
scale of robot datasets. The resulting video-action representation improves
generalization: it adapts from only 10--15 demonstrations, transfers across
embodiments without large-scale re-collection, and in some settings executes
new tasks zero-shot. At deployment time, sparse MoE inference, few-step
consistency distillation, Foresight Reasoning, and quantized execution allow
\method to perform real-time closed-loop control on real robots 
with a peak asynchronous execution frequency of 225 Hz. Together with a hierarchical VLM planner that decomposes long-horizon goals
into subtasks (\cref{sec:method}), these gains translate to more
consistent long-horizon behavior and high-precision manipulation, improving
over strong baselines such as $\pi_{0.5}$~\cite{pi2025pi05},
LingBot-VA~\cite{VA} by a substantial margin across
simulation and real-world evaluations.

In summary, our contributions are:
\begin{itemize}
  \item \textbf{Native video-action pretraining for robot control.} We pretrain
  the full video-action stack, including a semantic visual-action tokenizer and
  a causal DiT, using scalable self-supervision rather than adapting components
  pretrained for generic video generation.
  \item \textbf{A shared semantic video-action representation.} \method places
  world states and latent actions in a single semantic latent space anchored to
  a language-aligned visual foundation model, tightening
  vision--language--action alignment for stronger instruction following and
  action precision.
  \item \textbf{General, fast, and precise robot control.} The pretrained
  video-action representation supports few-shot and zero-shot generalization;
  sparse MoE inference, few-step distillation, Foresight Reasoning, and
  quantized deployment enable real-time closed-loop execution; and closed-loop re-grounding with
  hierarchical planning sustains fine-grained, long-horizon
  manipulation.
\end{itemize}

\section{Native Video-Action Model}
\label{sec:method}

\begin{figure*}[t]
  \centering
  \includegraphics[width=\linewidth]{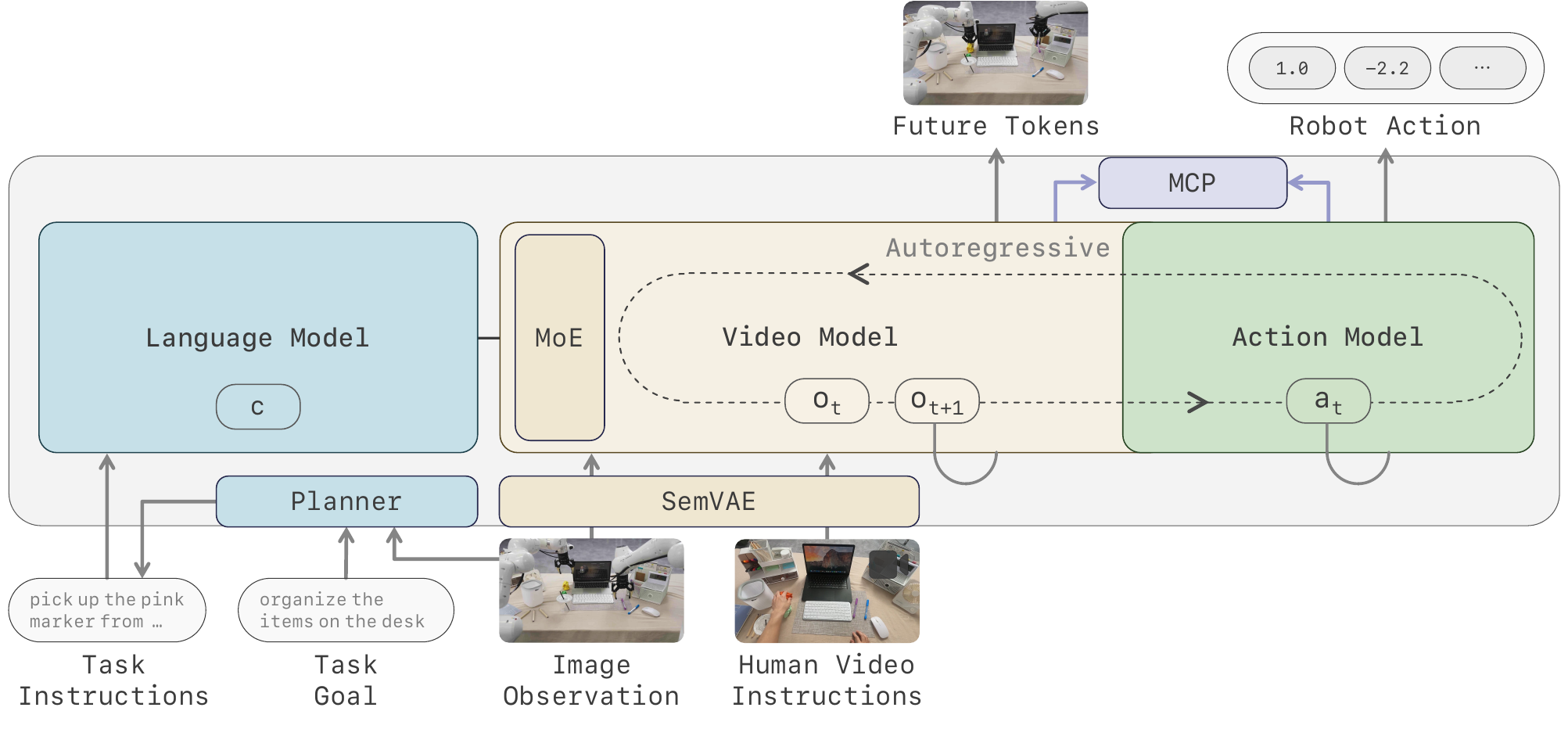}
  \caption{\textbf{Overview of \method.} The Planner
  (\cref{ssec:planning}) decomposes long-horizon goals into structured subtask
  context. The semantic visual-action tokenizer (\cref{subsec:tokenizer}) encodes
  image observations and human video prompts (\cref{ssec:incontext}) as visual
  latents. The video model with the MoE (\cref{ssec:architecture}) and the action
  model autoregressively predict future visual latents and robot actions. MCP
  (\cref{ssec:mcp}) adds future-chunk
  supervision, and the autoregressive rollout connects to asynchronous execution for real-time robot control.
  (\cref{ssec:foresight}).}
  \label{fig:framework}
\end{figure*}

This section presents the full \method stack (\cref{fig:framework}). After
reviewing how prior video-action models repurpose generic video generators for
control (\cref{subsec:preliminary}), we build the stack natively in two stages.
The first stage trains a semantic visual-action tokenizer, which aligns visual
latents with a frozen foundation model and extracts latent actions from
unlabeled video, yielding a shared visual-action latent space
(\cref{subsec:tokenizer}). The second stage pretrains a causal video-action
model on this space (\cref{subsec:pretraining}): a high-level VLM
\emph{planner} decomposes long-horizon goals into structured subtask context
(\cref{ssec:planning}); the low-level policy scales its video stream with a
sparse \emph{Mixture-of-Experts} DiT (\cref{ssec:architecture});
\emph{multi-chunk prediction} supervises multiple future chunks so the
representation captures trajectory-level dynamics (\cref{ssec:mcp});
\emph{in-context learning} admits human demonstration videos as visual task
prompts (\cref{ssec:incontext}); and human--robot co-training brings egocentric
human videos into the shared world model (\cref{ssec:egocentric}), under the
multi-task recipe of \cref{ssec:recipe}. At inference time, \emph{Foresight
Reasoning} overlaps prediction with execution and re-grounds each rollout on
the latest real observation (\cref{ssec:foresight}); post-training further
distills both experts into few-step samplers and applies inference acceleration
for real-time control (\cref{subsec:posttraining}).

\subsection{Preliminary}\label{subsec:preliminary}

We briefly review the video generative models that recent video-action models are
built upon, together with the formulation by which they are repurposed for
control.

\subsubsection{Video Generative Models}\label{ssec:video_generative_models}
Modern video generators comprise two components. First, a variational
autoencoder (VAE) $(\mathcal{E}, \mathcal{D})$ compresses a raw video
$o \in \mathbb{R}^{T \times H \times W \times 3}$ into a compact latent and
reconstructs it,
\begin{equation}
z = \mathcal{E}(o) \in \mathbb{R}^{T' \times h \times w \times c}, \qquad
\hat{o} = \mathcal{D}(z) \approx o,
\end{equation}
with temporal and spatial downsampling factors $f_t = T/T'$ and
$f_s = H/h = W/w$. 
Crucially, this VAE is trained purely for video compression: its
objective is reconstruction fidelity, with no notion of actions or
control. Second, a video generator---typically a flow-matching
transformer~\cite{flowmatching,cfm}---is trained in this latent space to
synthesize $z$ from noise $\epsilon \sim \mathcal{N}(0, I)$ conditioned on a
text prompt $c_{\text{text}}$. Concretely, it regresses the velocity field
along the interpolation path $z^{(s)} = (1-s)\epsilon + s\,z$:
\begin{equation}
\mathcal{L}_{\text{gen}} = \mathbb{E}_{s, \epsilon, z}\!\left[\,
\big\| v_\theta\big(z^{(s)}, s \mid c_{\text{text}}\big) - (z - \epsilon) \big\|^2 \,\right],
\label{eq:gen}
\end{equation}
where the attention is bidirectional over all $T'$ latent
frames~\cite{sora,wan2024video,veo}.

\subsubsection{Repurposing Video Generative Models for Robot Control}
Prior video-action models~\cite{VA,dreamzero} adapt this pretrained stack to
robotics through \emph{continued training} on robot data. Because the video
tokenizer temporally downsamples the observation stream while robot actions are
recorded at the original control frequency, one latent transition corresponds to
a block of $f_t$ low-level actions. We index the $T'$ visual latents as $z_{0:N}$ with
$N=T'-1$, and write the raw action sequence as $u_{0:f_t N-1}$. We group the raw
actions between two consecutive visual latents into an action chunk
\begin{equation}
a_t = u_{t f_t:(t+1)f_t-1}, \qquad t=0,\dots,N-1,
\label{eq:action-chunk}
\end{equation}
aligned with the transition $z_t \to z_{t+1}$. Throughout the video-action
model, $a_t$ denotes this action chunk rather than an individual low-level
control command. Two modifications are made. \emph{(i)~Video Causality.}
The bidirectional attention of \cref{eq:gen} is replaced by a causal
mask, so that each latent frame depends only on the past, matching closed-loop
control where the present cannot attend to the future:
\begin{equation}
p_\theta(z_{1:N} \mid z_0) =
\prod_{t=0}^{N-1} p_\theta\big(z_{t+1} \mid z_{\le t}\big).
\label{eq:causal}
\end{equation}
\emph{(ii)~Action injection.} A video-action model extends the video-only
factorization of \cref{eq:causal} to a conditional distribution over future
latents and actions given the initial visual context:
\begin{equation}
p_\theta\big(z_{1:N}, a_{0:N-1} \mid z_0\big) = \prod_{t=0}^{N-1}
\underbrace{p_\theta\big(z_{t+1} \mid z_{\le t}, a_{<t}\big)}_{\text{video generation:}\;\, (z_{\le t},\, a_{<t}) \,\to\, z_{t+1}}\;
\underbrace{p_\theta\big(a_t \mid z_{\le t}, a_{<t}, z_{t+1}\big)}_{\text{inverse dynamics:}\;\, (z_{\le t},\, a_{<t},\, z_{t+1}) \,\to\, a_t}.
\label{eq:va}
\end{equation}
Existing methods differ in how they parameterize \cref{eq:va}.
Joint prediction, as in DreamZero~\cite{dreamzero}, concatenates video and
action tokens and denoises each transition in a single shared stream. To make
the transition supervision explicit, we write the video-action objective as
separate video and action flow-matching losses. Let
$z_{t+1}^{(s)}=(1-s)\epsilon^z_t+s z_{t+1}$ and
$a_t^{(s)}=(1-s)\epsilon^a_t+s a_t$ denote their interpolated noisy targets:
\begin{align}
\mathcal{L}_{\text{vid}}
&= \mathbb{E}_{t,s,\epsilon^z_t}\!\left[
\big\| v^{\text{vid}}_\theta\big(z_{t+1}^{(s)}, s \mid z_{\le t}, a_{<t}\big)
- \big(z_{t+1}-\epsilon^z_t\big) \big\|^2 \right], \label{eq:vid-act-loss} \\
\mathcal{L}_{\text{act}}
&= \mathbb{E}_{t,s,\epsilon^a_t}\!\left[
\big\| v^{\text{act}}_\theta\big(a_t^{(s)}, s \mid z_{\le t}, a_{<t}, z_{t+1}\big)
- \big(a_t-\epsilon^a_t\big) \big\|^2 \right].
\end{align}
The complete video-action objective is
$\mathcal{L}_{\text{VA}}=\mathcal{L}_{\text{vid}}+\lambda_{\text{act}}\mathcal{L}_{\text{act}}$.
The Mixture-of-Transformers (MoT) of \methodold~\cite{VA,mot}
assigns the two modalities to separate expert networks
$\{v^{\text{vid}}_\theta, v^{\text{act}}_\theta\}$ that share a causal attention:
the video expert predicts the next visual latent and the action expert decodes
the action by inverse dynamics over the predicted transition,
\begin{equation}
\hat{z}_{t+1} = v^{\text{vid}}_\theta(z_{\le t}, a_{<t}), \qquad
\hat{a}_t = v^{\text{act}}_\theta\big(z_{\le t}, a_{<t}, \hat{z}_{t+1}\big).
\label{eq:mot}
\end{equation}
With a slight abuse of notation, $v^{\text{vid}}_\theta$ and $v^{\text{act}}_\theta$
denote the latent and action obtained by integrating their respective
flow-matching velocity fields, rather than the velocity fields themselves.

\subsection{Semantic Visual-Action Tokenizer}\label{subsec:tokenizer}

The compression-only VAE $(\mathcal{E}, \mathcal{D})$ of \cref{ssec:video_generative_models}
is a standard pixel-reconstruction tokenizer~\cite{kingma2013auto,wan2024video}:
its latents are optimized for reconstruction alone, without semantic alignment or
action supervision. We therefore follow RepWAM~\cite{repwam} and build our
tokenizer as the first stage of \method, augmenting reconstruction with two objectives: (1)~a semantic alignment
objective that pulls visual latents toward foundation-model features, and (2)~a
latent-action objective that extracts compact transition variables from
consecutive latents.

\subsubsection{Visual Tokenization with Semantic Alignment}
\begin{figure}[t]
  \centering
  \includegraphics[width=\linewidth]{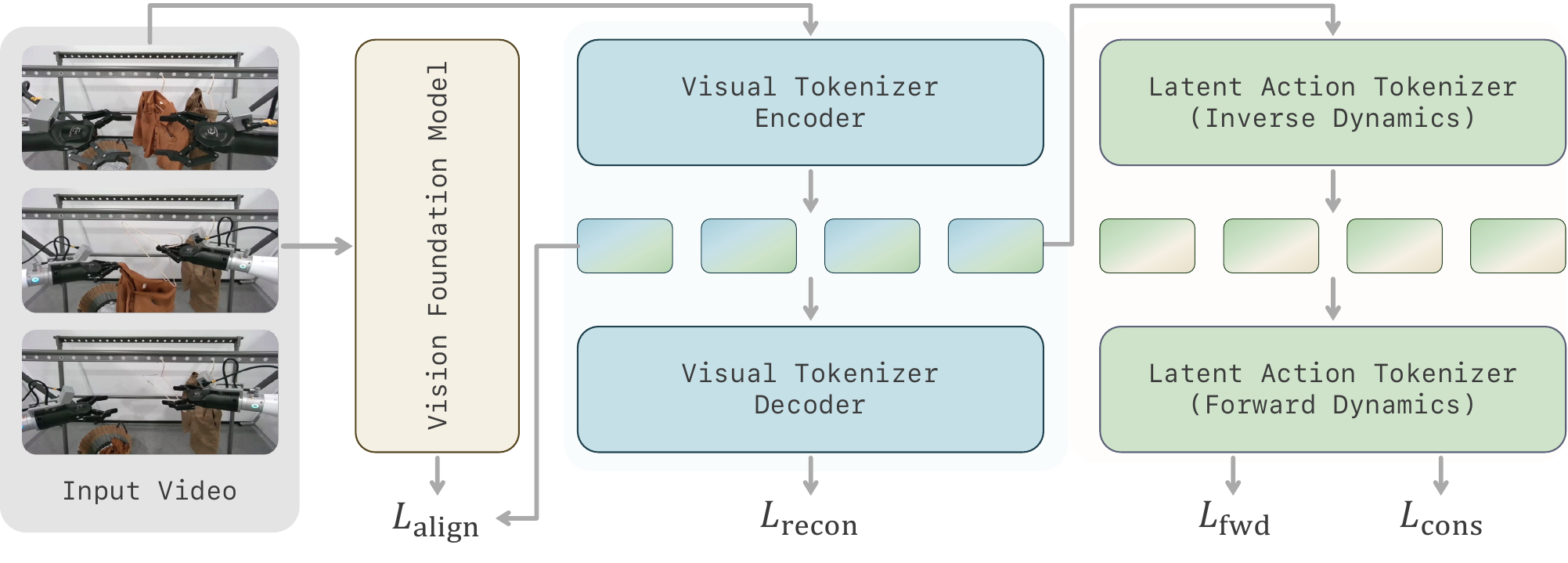}
  \caption{\textbf{Semantic visual-action tokenizer}. The visual tokenizer aligns
  reconstruction-oriented latents with visual foundation-model features, while
  the latent action tokenizer learns compact transition variables through
  inverse and forward dynamics. Adapted from RepWAM~\cite{repwam}.}
  \label{fig:tokenizer}
\end{figure}
We adopt a Vision Transformer (ViT) autoencoder for visual
tokenization~\cite{dosovitskiy2021an,wang2024omnitokenizer}. Given a video clip
$o_{1:T}$, our tokenizer encodes the clip in two parts. The first frame is encoded as
$16{\times}16$ spatial patches, and the subsequent frames are encoded as
$4{\times}16{\times}16$ spatiotemporal tubelets. The resulting tokens are
concatenated along the sequence dimension and passed to the encoder
$\mathcal{E}$, which outputs visual latents $z=\mathcal{E}(o)$. Our encoder uses
full spatial attention within each frame and causal attention across frames. A
symmetric decoder $\mathcal{D}$ then decodes the latents back to pixel space, $\hat{o}=\mathcal{D}(z)$.

The reconstruction loss combines pixel, perceptual, and adversarial terms:
\begin{equation}
\mathcal{L}_{\text{rec}} = \lambda_1 \| o - \hat{o} \|_1
+ \lambda_{\text{perc}}\,\mathcal{L}_{\text{perc}}(o, \hat{o})
+ \lambda_{\text{gan}}\,\mathcal{L}_{\text{gan}}(\hat{o}).
\end{equation}
For semantic alignment, we use a frozen Perception
Encoder~\cite{bolya2026perception} as the teacher model $G$. Since the tokenizer
latent and teacher feature may have different dimensions, a learnable projection
$W_{\text{align}}$ maps $z$ to the teacher feature dimension. We then match the
temporally pooled representations:
\begin{equation}
\mathcal{L}_{\text{align}} = \big\| \operatorname{avg}(W_{\text{align}}\, z)
- \operatorname{avg}\big(G(o)\big) \big\|_2^2 ,
\end{equation}
where $\operatorname{avg}$ denotes temporal average pooling. This
matches the clip-level semantics of $G$ while preserving per-frame information
for reconstruction. Overall, the visual tokenization objective is:
\begin{equation}
\mathcal{L}_{\text{vis}} = \mathcal{L}_{\text{rec}}
+ \lambda_{\text{align}}\,\mathcal{L}_{\text{align}} ,
\end{equation}
where $\lambda_{\text{align}}$ controls the strength of semantic alignment.

\subsubsection{Latent Action Tokenization}
We next build a latent action tokenizer upon these semantic visual latents
without using action labels. Following latent action models that infer actions
from observation transitions~\cite{genie,univla,moto}, each latent action is
defined as a compact transformation between two visual latents. We freeze the
visual tokenizer and train an inverse dynamics model (IDM) $q_\phi$ together
with a forward dynamics model (FDM) $f_\psi$. For two consecutive visual
latents, the IDM predicts:
\begin{equation}
\ell_t = q_\phi(z_t, z_{t+1}) \in \mathbb{R}^{d_\ell}, \qquad d_\ell \ll \dim(z_t) .
\end{equation}
The FDM decodes $\ell_t$ into a transport map and a residual, and reconstructs
the next latent:
\begin{equation}
\hat{z}_{t+1} = f_\psi(z_t, \ell_t) = K_t\, z_t + \delta_t ,
\end{equation}
where $(K_t, \delta_t)$ are predicted from $\ell_t$. The transport $K_t$ moves
information across spatial latent tokens, while $\delta_t$ accounts for changes
that cannot be explained by transport alone. A reverse transport
$\bar{f}_\psi$ predicts the previous latent from $(z_{t+1}, \ell_t)$ as
$\hat{z}_t = \bar{f}_\psi(z_{t+1}, \ell_t)$. Training uses forward prediction
and backward consistency on unlabeled video:
\begin{equation}
\mathcal{L}_{\ell} = \sum_t \big\| \hat{z}_{t+1} - z_{t+1} \big\|_2^2
+ \big\| \hat{z}_t - z_t \big\|_2^2 .
\end{equation}
The bottleneck on $\ell_t$ prevents it from copying the full visual state and
encourages it to capture control-relevant changes. It therefore provides an
action latent for modeling the transition from $z_t$ to $z_{t+1}$ in the
video-action model. Our tokenizer thus returns paired visual-action latents
$(z_{0:N}, \ell_{0:N-1})$ for each video clip, which serve as the training
targets for the second-stage video-action model, in which we write the learned
latent action token as $a_t \equiv \ell_t$ to match the notation in \cref{eq:va}.

\subsection{Video-action Model Pretraining and Inference}\label{subsec:pretraining}

Building on the semantic visual-action tokenizer, the second stage pretrains a
causal video-action model on the shared video-action latent space. We organize it
into two levels: a high-level \emph{planner} that decomposes the language goal
into subtasks and tracks their progress, and a low-level \emph{video-action
policy} that realizes each subtask by predicting future visual latents and the
latent actions connecting them. \textbf{The policy
operates at the granularity of \emph{chunks}, so throughout this subsection the
index $t$ ranges over chunks rather than individual frames.}

\subsubsection{Hierarchical Planning}\label{ssec:planning}
\begin{figure}[t]
  \centering
  \includegraphics[width=\linewidth]{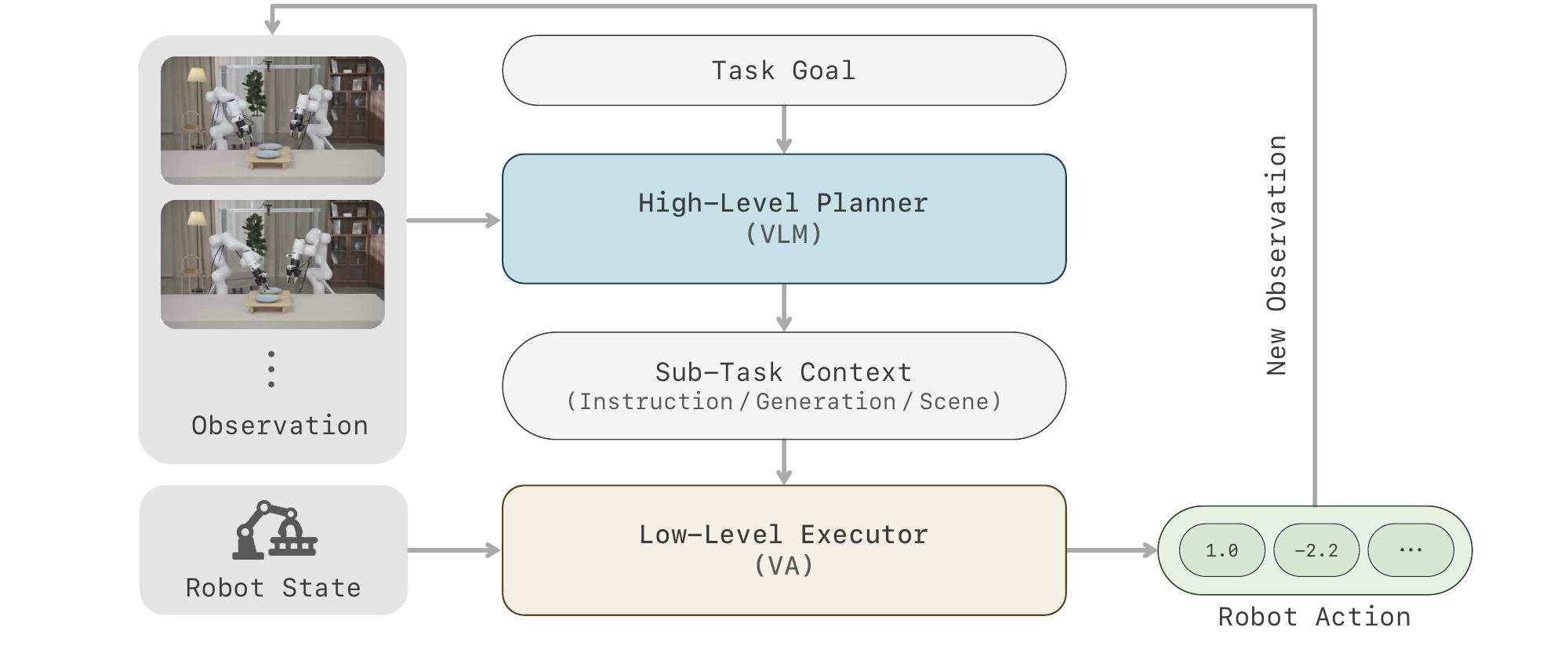}
  \caption{\textbf{Dual-system hierarchical policy}. A high-level VLM planner consumes
  the task goal and sparse recent observations to produce structured subtask
  context, which conditions the low-level video-action policy together with robot state
  to produce closed-loop actions.}
  \label{fig:planner}
\end{figure}
While the video-action policy handles continuous control within a single
subtask, long-horizon tasks require decomposing a high-level goal into subtasks,
tracking which subtask is currently active, and deciding when to transition to
the next one. We implement the planner as a pretrained vision-language model
(VLM) that is fine-tuned to produce structured subtask context for the policy.
The VLM backbone provides visual-language grounding out of the box, allowing the
planner to assess task state from sparse visual observations without learning
scene understanding from scratch.

The planner and the policy run at different frequencies and are decoupled
through an asynchronous shared buffer. The planner runs in the background at
${\sim}2$\,Hz and writes structured subtask context into the buffer; the policy
reads the latest context at each action-chunk boundary as its conditioning
signal. This ensures that planner inference latency does not block the policy's
execution loop.

\paragraph{Planner Output Schema.}
The planner's output format must exactly match the conditioning fields that the
video-action policy sees during training; any mismatch causes the policy to receive
out-of-distribution conditioning at inference time. Under this constraint, we
define the planner output as a structured JSON with four fields:
\begin{itemize}
  \item \textbf{\texttt{done}}: a completion signal indicating whether the
  current subtask should continue or the planner should transition to the next
  one.
  \item \textbf{\texttt{instruction}}: a compact execution directive for the
  current or next subtask, e.g., ``Pick up the red bottle on the table.'' This is
  the \emph{primary conditioning field} consumed by the policy during training.
  \item \textbf{\texttt{generation\_instruction}}: a richer description of the
  expected robot motion and object interaction within the subtask, giving the
  policy a more specific execution cue.
  \item \textbf{\texttt{local\_scene\_description}}: an observation-grounded
  description of the spatial layout and object state, providing local context for
  the policy.
\end{itemize}
The \texttt{done} field is used only by the high-level scheduler to decide
whether to keep or refresh the current subtask. The low-level policy is
conditioned on the three textual fields: \texttt{instruction} provides the
compact command, while \texttt{generation\_instruction} and
\texttt{local\_scene\_description} provide more detailed action and scene
context when available.

\paragraph{Planner Data Construction.}
Training data for the planner is constructed from robot demonstration videos
with segment-level annotations. Each video contains an episode-level task goal
and a temporal sequence of segments $[g_0, g_1, \ldots, g_N]$, where each
segment is annotated with the four fields above.

The core construction method is \emph{boundary-crossing sampling}, designed to
directly align with the planner's online deployment scenario. At each segment
boundary, we generate two types of training samples:
\begin{itemize}
  \item \textbf{done\,=\,false}: the observation time $t^*$ falls inside the
  current segment $g_i$; the target is the four fields of $g_i$ (echo the
  current subtask).
  \item \textbf{done\,=\,true}: the observation time $t^*$ has just crossed the
  $g_i \to g_{i+1}$ boundary; the target is the four fields of $g_{i+1}$
  (predict the next step).
\end{itemize}
Each sample's input consists of:
(i)~three keyframes at $t^*{-}2$\,s, $t^*{-}1$\,s, and $t^*$, providing
short-term temporal context;
(ii)~the episode-level task goal;
(iii)~the full text history of all completed prior segments $g_0, \ldots,
g_{i-1}$;
and (iv)~the current subtask description (for done\,=\,false samples only).
The \texttt{done\,=\,true} samples thus require the planner to predict the
upcoming subtask from the visual observation and task history, rather than
merely echoing the current one.

The training set is task-balanced to prevent high-frequency tasks from
dominating.

\paragraph{Planner Training.}
The planner is trained by LoRA fine-tuning a pretrained VLM backbone with a
frozen vision tower. This keeps the pretrained visual grounding ability intact
while adapting the language head to the structured subtask schema. The planner
is therefore treated as an interface module that supplies local language context
to the video-action policy, rather than as a separately optimized policy
component.

\subsubsection{Architecture: Mixture-of-Experts Video Stream}\label{ssec:architecture}
The low-level video-action policy keeps the Mixture-of-Transformers (MoT)
structure of \cref{eq:mot}: the video and action modalities are handled by
two experts $\{v^{\text{vid}}_\theta, v^{\text{act}}_\theta\}$ that share one
causal self-attention but own separate feed-forward pathways. Because visual
dynamics carry most of the modeling burden, we scale the two streams
asymmetrically: the \emph{video expert} replaces the dense feed-forward network
in each causal DiT block with a sparse Mixture-of-Experts (MoE) routed layer,
while the \emph{action expert} remains dense. The MoE design follows the sparse
scaling principle of DeepSeekMoE and DeepSeek-V3~\cite{deepseekmoe,deepseekv3},
instantiated inside our causal video-action DiT. Self-attention remains causal
over the visual-action history and cross-attention remains conditioned on
language. Since routing is applied after causal self-attention as a token-wise
feed-forward operation, increasing the number of experts increases the video
stream's capacity while preserving the same causal dependency structure
described in \cref{eq:va}.

For a hidden token $h \in \mathbb{R}^{d}$, the router uses a gate matrix
$W_g=[g_1,\dots,g_{N_e}]^\top \in \mathbb{R}^{N_e \times d}$ to compute expert
scores $r(h) \in \mathbb{R}^{N_e}$,
\begin{equation}
r_i(h) = \sigma(g_i^\top h), \qquad i=1,\dots,N_e ,
\end{equation}
where $\sigma(\cdot)$ is the sigmoid function and $N_e$ is the number of routed
experts. We use group-limited top-$k$
routing. Experts are partitioned into groups; a small number of groups are first
selected by their strongest router scores, and the final $k$ experts are then
chosen within those groups. The selected set is
\begin{equation}
\mathcal{R}(h) =
\operatorname{TopK}_{\mathrm{group}}\big(r(h)+b, k\big), \qquad
\mathcal{R}(h) \subset \{1,\dots,N_e\},\; |\mathcal{R}(h)|=k ,
\end{equation}
where $b \in \mathbb{R}^{N_e}$ is an expert-wise correction bias used only for
expert selection. The top-$k$ scores are normalized and scaled as
\begin{equation}
\alpha_i(h) =
\gamma \frac{r_i(h)}{\sum_{j \in \mathcal{R}(h)} r_j(h)}, \qquad
i \in \mathcal{R}(h),
\end{equation}
where $\alpha_i(h)$ is a scalar routing weight and $\gamma$ is a routed scaling
factor. The MoE output is then
\begin{equation}
\operatorname{MoE}(h) =
E_{\mathrm{shared}}(h)
+ \sum_{i \in \mathcal{R}(h)} \alpha_i(h) E_i(h),
\end{equation}
where $E_{\mathrm{shared}}:\mathbb{R}^{d}\to\mathbb{R}^{d}$ captures common
transformations shared by all tokens, and
$E_i:\mathbb{R}^{d}\to\mathbb{R}^{d}$ are routed SwiGLU experts. In our main
configuration, $N_e{=}128$ and $k{=}8$, so each token activates only a small
fraction of the routed expert pool while the full parameter set remains
available across the sequence.

Stable MoE training requires balanced expert utilization. We therefore use the
auxiliary-loss-free load-balancing strategy of Loss-Free
Balancing~\cite{lossfreebalance}, also adopted in DeepSeek-V3~\cite{deepseekv3},
and follow the auxfree bias update used in Moonlight~\cite{moonlight}.
After each optimization step, the correction bias is updated from the observed
token count $c_i$ assigned to each expert,
\begin{equation}
v_i = \operatorname{sign}(c_i-\bar{c}), \qquad
b_i \leftarrow b_i - \eta_{\mathrm{lb}}
\left(v_i-\frac{1}{N_e}\sum_{j=1}^{N_e}v_j\right),
\end{equation}
where $\bar{c}=\frac{1}{N_e}\sum_{i=1}^{N_e}c_i$ is the mean expert load and
$\eta_{\mathrm{lb}}$ is the balance update rate. This discourages overloaded
experts and promotes underused experts without injecting large balancing
gradients into the video-action diffusion objective. We keep a lightweight
sequence-wise router regularizer and expert-utilization statistics for
monitoring. The forward computation is executed with grouped expert matrix
multiplications for efficient long-sequence training.

\begin{figure}[t]
    \centering
    \includegraphics[width=\linewidth]{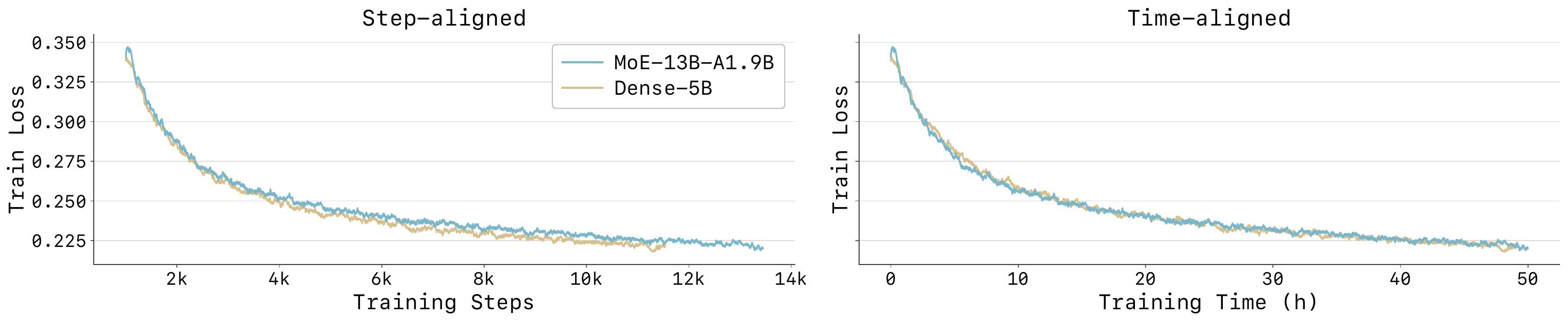}
    \caption{
    \textbf{Training loss comparison between the MoE-13B-A1.9B video-stream model and
    the Dense-5B baseline.} Left: loss aligned by optimization steps, where both
    models remain close and the dense baseline is slightly lower. Right: loss
    aligned by wall-clock training time, where the final losses nearly coincide.
    }
    \label{fig:moe_dense_loss}
\end{figure}

As a final check of the optimization behavior of this sparse video-stream
design, we compare the training loss of the MoE-13B-A1.9B model with a
Dense-5B baseline under the same training setup. As shown in
\cref{fig:moe_dense_loss}, when compared at the same number of optimization
steps, the two models follow very similar loss trajectories, with the dense
baseline being slightly lower. When compared at matched wall-clock training
time, their final losses nearly coincide. This indicates that the proposed MoE
video stream increases the total model capacity without introducing a noticeable
optimization penalty under the considered training budget.

\subsubsection{Multi-chunk Prediction}\label{ssec:mcp} 

\begin{figure}[t]
  \centering
  \includegraphics[width=\linewidth]{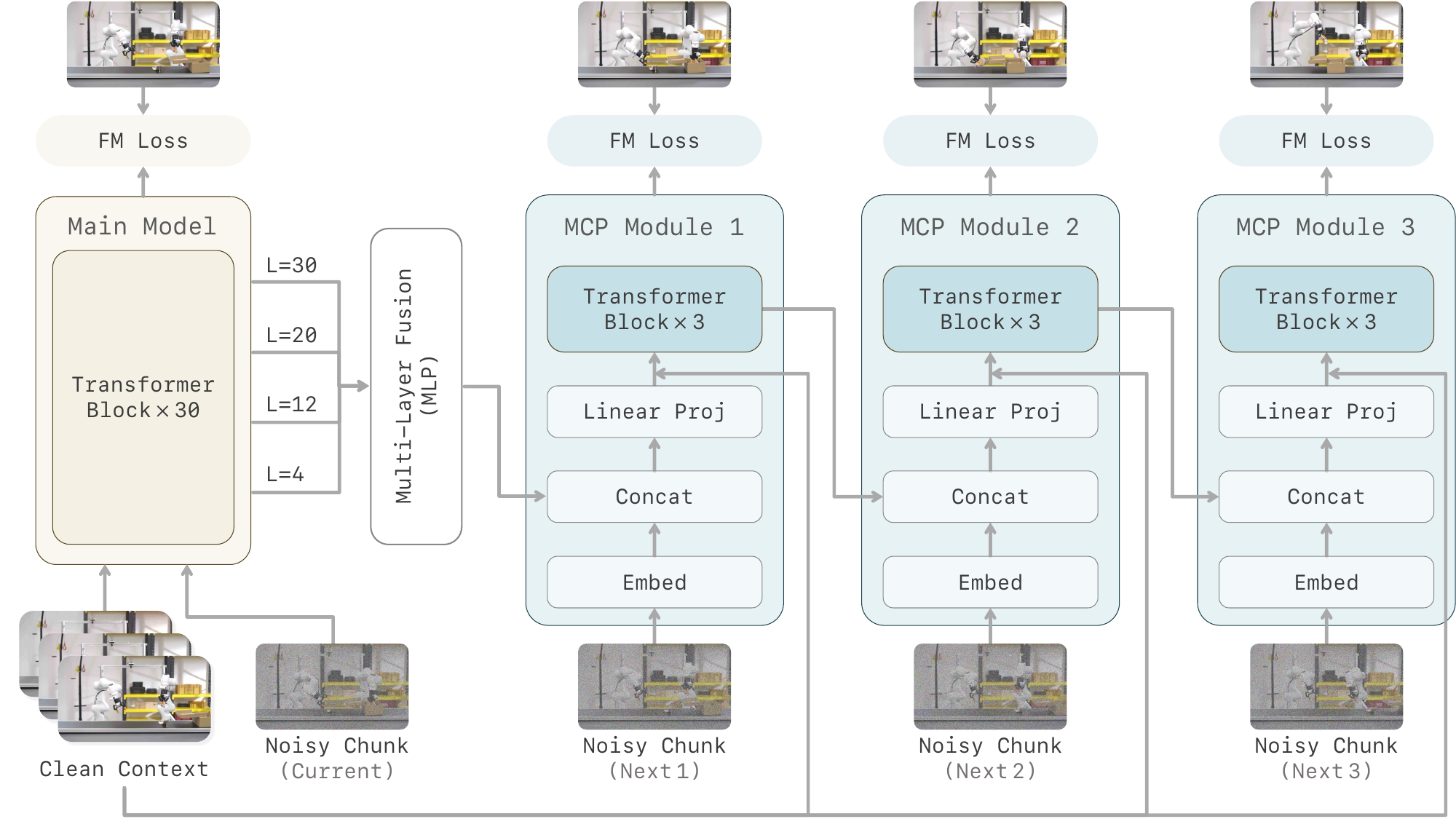}
  \caption{\textbf{Multi-Chunk Prediction (MCP)}. Lightweight auxiliary modules predict
  several future latent chunks at increasing horizons from the main model's
  representation, providing dense temporal supervision during training. Adapted
  from Next Forcing~\cite{nextforcing}.}
  \label{fig:mcp}
\end{figure}

Following \methodold~\cite{VA}, we train the video-action model with teacher forcing, predicting each chunk from the ground-truth history. However, standard teacher-forced video-action training supervises the model only on the current chunk, creating a \emph{myopic supervision} problem in which the learning signal becomes overly local: adjacent chunks, especially at high frame rates, are often visually similar enough that the model can reduce loss by copying appearance rather than learning the underlying dynamics.

This is misaligned with the goal of \method, which is to pretrain a native video-action foundation model whose representations capture how the physical world evolves under actions, rather than only preserving short-term visual continuity. Building on the semantic visual-action latent space produced by our tokenizer, we therefore adopt Multi-Chunk Prediction (MCP), inspired by Next Forcing~\cite{nextforcing}, as an auxiliary training objective that makes the causal video-action policy's visual representations predictive of future state transitions.

Teacher forcing supervises only the next-chunk conditional of the visual stream,
\begin{equation}
p_\theta\big(z_{t+1} \mid z_{\le t}, a_{<t}\big),
\label{eq:tf}
\end{equation}
matching each prediction against the immediate next chunk under the ground-truth
visual-action prefix $(z_{\le t},a_{<t})$. MCP instead predicts the whole block
of the next $K$ chunks from this context, supervising horizon-wise future
conditionals
\begin{equation}
p_\theta\big(z_{t+1:t+K} \mid z_{\le t}, a_{<t}\big)
\approx \prod_{k=1}^{K}
p_{\theta,k}\big(z_{t+k} \mid h_t^{(k-1)}\big),
\qquad h_t^{(0)} = f_\theta(z_{\le t},a_{<t}),
\label{eq:mcp-cond}
\end{equation}
so the representation at step $t$ must encode how the scene evolves over the next
$K$ chunks, not just the next one. Here $f_\theta$ denotes the main DiT's
feature extraction over the ground-truth prefix, and $h_t^{(k-1)}$ is the
internal feature state passed through the stacked MCP modules, not a clean or
predicted future observation. Thus the predictor for horizon $t+k$ reuses previous MCP features
rather than conditioning on $z_{t+1:t+k-1}$. Each horizon $k$ is trained with the
same flow-matching objective as the main model. Specifically, with
$z_{t+k}^{(s)}=(1-s)\epsilon^k_t+s z_{t+k}$, the MCP loss is
\begin{equation}
\mathcal{L}^{\text{MCP}}_k
= \mathbb{E}_{t,s,\epsilon^k_t}\!\left[
\big\| v^{\text{MCP}}_{\theta,k}\big(z_{t+k}^{(s)}, s \mid h_t^{(k-1)}\big)
- \big(z_{t+k}-\epsilon^k_t\big) \big\|^2 \right],
\label{eq:mcp-loss}
\end{equation}
where $v^{\text{MCP}}_{\theta,k}$ is the velocity field of the $k$-th MCP
module. 

We realize each conditional in \cref{eq:mcp-cond} with a lightweight MCP module
attached to the main DiT (\cref{fig:mcp}), one per horizon. Following
Next Forcing~\cite{nextforcing}, we use three future chunks
($K=3$; next$^1$, next$^2$, next$^3$). Each
module denoises a temporally shifted visual-latent target under the same
flow-matching formulation, but with a larger timestep shift, so it cannot solve
the task from its own noisy input and must instead rely on the main model's
representation. The modules form a causal chain: the first consumes fused
intermediate features from several DiT layers, and each deeper module additionally
conditions on the previous one, letting near-future predictions inform
farther-future ones. Gradients from future-chunk denoising thus flow back into
the backbone as dense temporal supervision, pushing it to organize its latents
around trajectory-level dynamics rather than local appearance. MCP acts only on
the visual stream, but the more predictive visual states also improve action
decoding through the shared video-action attention and inverse-dynamics pathway.

The MCP auxiliary objective combines the horizon losses with the same weights as
Next Forcing, $(w_1,w_2,w_3)=(0.5,0.2,0.1)$:
\begin{equation}
  \mathcal{L}_{\text{MCP}} = \sum_{k=1}^{K} w_k \mathcal{L}^{\text{MCP}}_k .
  \label{eq:mcp-obj}
\end{equation}
At deployment, the same trained checkpoint supports two modes: the MCP modules can be discarded for zero-overhead inference, leaving the main policy improved purely by training-time supervision, or retained to predict the next visual chunk in parallel with the current one, reducing rollout latency for closed-loop control.

\subsubsection{In-context Learning}\label{ssec:incontext}
Existing video-action models rely on language instructions as the primary task specification. The robot receives a textual command and is expected to infer the intended manipulation procedure from language alone. However, for rare-object manipulation, complex multi-object interaction, and long-horizon tasks, language descriptions are rarely exhaustive and may fail to specify the fine-grained temporal structure of the operation.

To provide a richer task specification, following Zero-WAM~\cite{zerowam} and DVA~\cite{rhoda_dva}, we introduce video in-context learning (ICL). In each ICL sample, the robot video is paired with a semantically aligned human demonstration video. The human video does not need to share the same object instance, viewpoint, or scene layout as the robot video. Instead, it serves as a dynamic visual example that demonstrates how the manipulation task should unfold. Compared with language-only instructions, such a visual context provides explicit temporal and procedural guidance, helping the model follow complex task semantics and generalize to unseen open-world tasks~\cite{zhou2025exploring}.

Given a human demonstration video, we encode it with our semantic tokenizer into an
in-context latent $z_{\mathrm{icl}}$. Following the teacher-forced
video-action pretraining objective in \cref{eq:va}, the ICL human video augments
robot prediction by conditioning every robot chunk on $z_{\mathrm{icl}}$. For a
paired robot trajectory with visual chunks $z_{0:N}$ and action chunks
$a_{0:N-1}$, the in-context video-action objective is:
\begin{equation}
p_\theta\big(z_{1:N}, a_{0:N-1} \mid z_0, z_{\mathrm{icl}}\big)
= \prod_{t=0}^{N-1}
p_\theta\big(z_{t+1} \mid z_{\le t}, a_{<t}, z_{\mathrm{icl}}\big)\,
p_\theta\big(a_t \mid z_{\le t}, a_{<t}, z_{t+1}, z_{\mathrm{icl}}\big).
\label{eq:causal_icl}
\end{equation}
Here, the in-context latent $z_{\mathrm{icl}}$ is treated as an external task
prompt and is available to all robot prediction steps, while the robot-side
causal mask prevents each chunk $z_t$ from attending to future robot chunks.
Since $z_{\mathrm{icl}}$ comes from a separate human demonstration rather than
the future of the robot trajectory, attending to the full demonstration does not
violate robot-side causality. The same conditioning signal guides both future
robot video prediction and the corresponding action prediction.

\subsubsection{Human-Robot Data Co-training}\label{ssec:egocentric}
Robot manipulation data is expensive to collect and limited in diversity, while
egocentric human manipulation videos are cheap to scale and cover a much broader
range of tasks and scenes. Since prior work shows that human data transfers to
robot policies, especially under scarce robot data~\cite{whatmatters,zhou2024mitigating,egomimic,mimicplay},
we \emph{co-train} on human and robot data together rather than pretraining on
human videos and adapting later~\cite{beingh0,beingh07}, allowing the shared
video-action model to improve real-robot control throughout training.

Naive joint training faces two gaps: an \emph{action-space mismatch} between
human hands and robot grippers, and a \emph{motion gap} between natural human
motion and robot kinematics~\cite{whatmatters}. These gaps are especially harmful
for a video-action \emph{world model}, where past actions feed back into the
prediction context (\cref{eq:va}) and misaligned human actions can corrupt the
shared dynamics. We address them by mapping human hand poses into the robot
action space while keeping embodiment-specific action encoders and decoders
around a shared world model. Following~\cite{whatmatters}, we additionally map
the hand to a functional gripper opening rather than a placeholder, so human
actions also supervise when to grasp.

We co-train two corpora that share the observation $o$ but differ in their native
actions: a robot set $\mathcal{D}_R=\{(o,a^R)\}$ and a human set
$\mathcal{D}_H=\{(o,\tilde a^H)\}$. All actions are expressed in a unified cross-embodiment action layout in which
channels missing for an embodiment are zero-padded and masked; its
$d_a$-dimensional bimanual end-effector slice $\mathcal{A}$ holds, for each
hand, a 6-DoF root pose together with a scalar gripper opening. Human hand
labels are retargeted into exactly this slice by a map $\Phi$, with the
remaining channels padded and masked as for any other embodiment:
\begin{equation}
a^H=\Phi(\tilde a^H)\in\mathcal{A}.
\label{eq:ego-retarget}
\end{equation}
Here $\Phi$ keeps the two 6-DoF hand-root poses and maps each hand's finger
configuration $q$ to a single scalar gripper opening $\varphi(q)$, populating
$\mathcal{A}$ exactly as a robot action does. This functional opening is what replaces the
no-op gripper placeholder used in prior co-training~\cite{whatmatters}. After
retargeting, the two domains share an \emph{identical} action representation; what
remains is the motion gap---the same tuple carries different distributions and
physical meaning for a human hand and a robot end-effector---which the per-domain
action heads absorb.

Both domains pass through one shared world model and differ only in a per-domain
action encoder $E_d$ and decoder $P_d$ ($d\in\{R,H\}$); the video expert
$v^{\text{vid}}$, the action expert $v^{\text{act}}$, and the causal attention of
\cref{eq:mot} are all shared:
\begin{equation}
\hat a^{(d)}_t = P_d\big(v^{\text{act}}(z_{\le t},\,E_d(a_{<t}),\,\hat z_{t+1})\big).
\label{eq:ego-splitio}
\end{equation}
The objective sums both domains, each decoded through its own action head:
\begin{equation}
\mathcal{L}_{\text{co-train}}
=\mathbb{E}_{\mathcal{D}_R}\!\big[\mathcal{L}_{\text{vid}}+\mathcal{L}^{R}_{\text{act}}\big]
+\mathbb{E}_{\mathcal{D}_H}\!\big[\mathcal{L}_{\text{vid}}+\mathcal{L}^{H}_{\text{act}}\big].
\label{eq:ego-cotrain}
\end{equation}
Here $\mathcal{L}_{\text{vid}}$ and $\mathcal{L}^{d}_{\text{act}}$ are the
model's flow-matching video and action losses, the action term decoded through the
domain head $P_d$. The shared video and action experts see both domains and learn
cross-domain dynamics from human data; the domain-specific heads $E_d,P_d$ only
isolate the low-level action parametrization of each embodiment from the shared
representation.
At deployment we use only the robot branch $(E_R,P_R)$, so co-training adds no
inference cost.

Concretely, the hand-to-gripper map $\varphi$ treats the hand as a virtual parallel gripper,
with the thumb as one jaw and the four fingers as the other. From the captured
finger joints we recover 3D fingertip positions by forward kinematics and take the
opening as the distance between the two sides along the closing direction,
yielding a metric aperture that is quantile-normalized per dataset in the same
way as the robot gripper channels. Measuring this over the whole finger envelope rather than
a single thumb-to-index pair keeps it stable when a finger is mistracked. The
6-DoF hand-root poses are kept as is, with no conversion. The corpus and data pipeline are described in
\cref{ssec:egodata}. During training, the robot and human action heads in
\cref{eq:ego-splitio} are selected according to each sample's embodiment, while
the shared video-action backbone is updated by both domains. We initialize the
human action heads from the robot heads, providing a stable starting point for
co-training.

\subsubsection{Training Recipe}\label{ssec:recipe}
Generic video generators such as Wan~\cite{wan2024video} are pretrained for
synthesis rather than embodied interaction, and their bidirectional attention
conflicts with the causal regime of closed-loop control (\cref{eq:causal});
combined with our purpose-built tokenizer, whose semantic latent space differs
from that of any off-the-shelf generator, this leads us to build the
pretraining recipe from scratch rather than continue-train a borrowed backbone.

Rather than training the video-action model in fully separate, sequential
stages, we adopt a \emph{multi-task co-training} recipe in which all objectives
are optimized jointly throughout pretraining and only their relative mixing
weights change over time. A strictly staged curriculum---first text-to-image,
then text-to-video, then video-action---risks catastrophically forgetting the
broad appearance and dynamics priors acquired earlier once training narrows to
scarce embodied data. Co-training instead keeps every objective present at all
times, so the model retains general world knowledge while progressively
specializing toward control. All five tasks are instantiated on the shared
semantic latent space of our tokenizer and the same causal DiT backbone, differing only
in their conditioning and supervision:
\begin{itemize}
  \item \textbf{T2I} (text-to-image): single-frame text-conditioned generation
  that builds visual-semantic grounding and aligns the latent space with
  language.
  \item \textbf{T2V} (text-to-video): text-conditioned multi-frame generation
  that learns general temporal dynamics and motion priors from web-scale video.
  \item \textbf{TI2VA} (text-image to video-action): the core control task,
  combining future-latent \emph{video prediction} with \emph{inverse-dynamics
  modeling} of the latent actions that connect consecutive states
  (\cref{eq:va}), conditioned on a language instruction and an observation
  history.
  \item \textbf{ICL} (in-context learning): given a demonstration video as
  additional context, the model generates the video-action rollout for a new
  instance of the same task, enabling few-shot adaptation without weight
  updates.
  \item \textbf{HCT} (human-robot co-training): joint training on egocentric
  human videos and robot data through a shared world model with
  embodiment-specific action encoders and decoders (\cref{ssec:egocentric}),
  transferring low-cost, diverse human demonstrations to robot control.
\end{itemize}
All tasks share parameters and are optimized jointly under a single mixed
objective. Let $\mathcal{T}$ denote the task set containing T2I, T2V, TI2VA,
ICL, and HCT, and let each task $i\in\mathcal{T}$ contribute a loss
$\mathcal{L}_i$ on the shared latent space and DiT backbone. At training progress
$\tau\in[0,1]$ we sample tasks from a distribution $\pi(\tau)$ over
$\mathcal{T}$ and minimize the weighted objective
\begin{equation}
\mathcal{L}(\tau) = \sum_{i\in\mathcal{T}} \pi_i(\tau)\,\mathcal{L}_i,
\qquad \pi_i(\tau)\ge 0,\quad \sum_{i\in\mathcal{T}}\pi_i(\tau)=1 .
\label{eq:recipe}
\end{equation}
Here $\mathcal{L}_{\text{T2I}}$ and $\mathcal{L}_{\text{T2V}}$ are flow-matching
generation losses (\cref{eq:gen}); $\mathcal{L}_{\text{TI2VA}}$ combines the
video-prediction and inverse-dynamics terms of \cref{eq:va} with the multi-chunk
objective $\mathcal{L}_{\text{MCP}}$ (\cref{eq:mcp-obj}); $\mathcal{L}_{\text{ICL}}$
is the in-context objective of \cref{eq:causal_icl}; and $\mathcal{L}_{\text{HCT}}$
is the human-robot co-training loss of \cref{ssec:egocentric}.

The sampling distribution $\pi(\tau)$ follows a \emph{coarse-to-fine} schedule.
Early training ($\tau\!\to\!0$) places most mass on T2I to establish appearance
and semantic alignment; the mass then shifts toward T2V to consolidate temporal
dynamics; and late training ($\tau\!\to\!1$) concentrates on the video-action
tasks (TI2VA, ICL, HCT) to specialize the shared representation for
action-relevant control. Because all tasks share parameters, earlier-emphasized
objectives still provide a small but nonzero regularizing signal in later phases,
preserving the broad image- and video-priors while the model adapts to embodied
control.

\subsubsection{Foresight Reasoning}\label{ssec:foresight}
\begin{figure}[t]
  \centering
  \includegraphics[width=\linewidth]{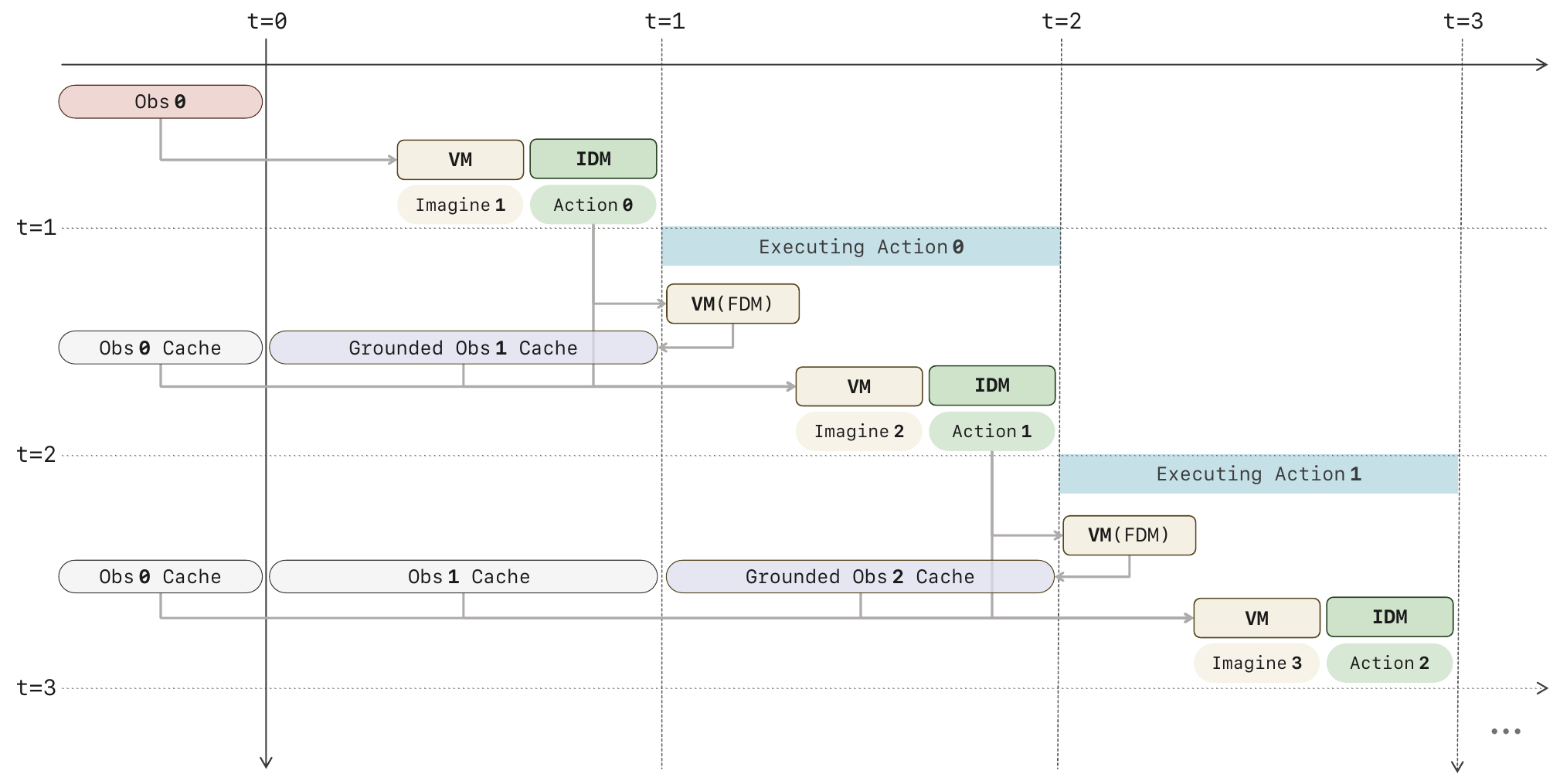}
  \caption{\textbf{Foresight Reasoning as an asynchronous inference/execution flow}. A
  cold start initializes the KV cache and pre-computes the first action chunk.
  During the loop, the executor runs the current chunk while real observations
  enter \textsc{ObsQueue}; the inference branch re-grounds the KV cache with
  those observations, appends the currently executing action, imagines its visual
  outcome with FDM grounding, and predicts the next action chunk before execution
  stalls.}
  \label{fig:foresight}
\end{figure}
In the multi-stream view of Multi-Stream LLMs~\cite{multistream}, a single
serial stream of computation is self-blocking: the model must finish computing
before the robot can act, and the robot must finish acting before a new
observation returns. We therefore organize the closed-loop rollout as two
causally coupled streams that advance at different rates: a \emph{prediction stream},
in which the model predicts future visual latents and the actions decoded from
them, and an \emph{execution stream}, in which the robot executes the current action
chunk in the real world. Run synchronously, the two streams stall each other:
after executing chunk $a_t$ the robot idles while the model denoises the next
chunk, turning model latency into control latency.

Foresight Reasoning runs the two streams \emph{asynchronously}. While the robot
executes action chunk $a_t$, the prediction stream prepares the next action
chunk $a_{t+1}$ before the real observation of $a_t$ has returned. It first
appends the currently executing action chunk $a_t$ to the latest
feedback-grounded cache $C_t$---the KV-cache entries summarizing the grounded
history $(z_{\le t}, a_{<t})$---giving the temporary context
$C_{\mathrm{tmp}}=C_t\cup\{a_t\}$. The policy's video expert then runs a
forward-dynamics pass, performed online by the policy itself rather than by the
tokenizer's frozen FDM $f_\psi$, that imagines the visual outcome of the current
chunk,
\begin{equation}
\hat{z}_{t+1}
= \mathrm{FDM}_\theta\big(C_t \cup \{a_t\}\big)
= v^{\text{vid}}_\theta\big(z_{\le t},\, a_{\le t}\big),
\label{eq:foresight-ground}
\end{equation}
and the action expert decodes the next action chunk from
$C_{\mathrm{tmp}}\cup\{\hat{z}_{t+1}\}$. Thus $a_{t+1}$ is ready when execution
of $a_t$ finishes, hiding prediction latency behind motion. Running ahead,
however, is not free: if the prediction stream keeps feeding its own
$\hat{z}_{t+1}$ back into the context, the rollout becomes open-loop; because the
video backbone favors temporal smoothness, it continues this hallucinated
trajectory, ignores physical feedback, and drifts until the policy can no longer
react.

We therefore re-ground the rollout whenever a real observation arrives. After
chunk $a_t$ completes, the returned observation is encoded into the true latent
$z_{t+1}$ and overwrites the stale predicted context $\hat{z}_{t+1}$ in the KV
cache, together with the executed action chunk $a_t$. The next prediction step
therefore starts from the feedback-grounded cache $C_{t+1}$ rather than from an
unverified imagined latent. The asynchronous rollout thus stays closed-loop---a
\emph{predict-then-correct} step in which the fast prediction stream drafts ahead
while each real observation pulls it back. To train the video expert for this
role, post-training adds a forward-dynamics grounding loss that supervises it to
predict the next visual latent from the feedback-grounded context and executed
action, following the flow-matching objective of \cref{eq:gen}:
\begin{equation}
\mathcal{L}_{\text{FDM}} = \mathbb{E}_{t,\,s,\,\epsilon}\!\left[\,
\big\| v^{\text{vid}}_\theta\big(z^{(s)}_{t+1}, s \mid z_{\le t},\, a_{\le t}\big)
- (z_{t+1} - \epsilon) \big\|^2 \,\right],
\label{eq:fdm}
\end{equation}
matching the conditioning of \cref{eq:foresight-ground} used during asynchronous
inference. Unlike the video loss $\mathcal{L}_{\text{vid}}$
(\cref{eq:vid-act-loss}), which conditions only on the past actions $a_{<t}$,
this objective additionally feeds the executed action $a_t$ (i.e.\ $a_{\le t}$),
turning the video expert into a genuine forward-dynamics predictor of the current
transition.
Unlike Multi-Stream
LLMs~\cite{multistream}, whose streams carry input/output roles, our streams carry
predicted future world states, so closed-loop control hinges on re-grounding them
against real observations---a step with no counterpart in the language setting.
Foresight Reasoning is an inference-time scheme independent of the MCP modules of
\cref{ssec:mcp}: the two can be enabled separately.

\paragraph{Relation to Leapfrog Inference.}
Direct Video-Action (DVA) proposes Leapfrog Inference as a related way to hide
model latency behind robot execution~\cite{rhoda_dva}. In both cases, the policy
does not wait for the current motion to finish before preparing the next one:
prediction advances while execution is still underway. The key difference is the
state carried by the asynchronous stream. Leapfrog Inference advances a causal
video-action rollout far enough into the future that the next action can be
decoded after the current action has completed. Foresight Reasoning instead
treats the predicted future latent as an FDM-grounded cache update. At each
chunk, the stream starts from the observation-grounded KV cache, conditions on
the currently executing action, and uses the policy's forward-dynamics model
(FDM) to predict the next visual latent. This predicted latent is inserted into
the cache so the action expert can decode the future action before execution
stalls. As soon as the real observation arrives, it re-grounds the cache by
overwriting the drafted latent, after which the next FDM prediction starts from
the corrected state. Thus our asynchronous execution is explicitly
action-conditioned and closed-loop at every chunk, while still hiding most of the
next-chunk prediction latency.

\subsection{Post-training}\label{subsec:posttraining}

\subsubsection{Diffusion Distillation}

The post-trained policy samples each chunk by integrating the flow-matching
velocity field along the probability-flow (PF) ODE; in \methodold this defaults to
$5$ denoising steps for the video expert and $10$ for the action expert. To make
the policy real-time, we distill it into a few-step \emph{consistency
model}~\cite{consistencymodel}. A consistency model $f_\xi(x^{(s)},s)$ maps any
point on a PF-ODE trajectory directly to its clean endpoint, satisfying the
self-consistency property $f_\xi(x^{(s)},s)=f_\xi(x^{(s')},s')$ for any
$s,s'$ on the same trajectory and the boundary condition
$f_\xi(x^{(1)},1)=x^{(1)}$ at the data end. To lighten notation, $x$, $v_\theta$,
and $f_\xi$ stand for either expert---the video expert acting on visual latents
$z$ or the action expert acting on action chunks $a$---each retaining its usual
causal context conditioning $(z_{\le t}, a_{<t}, \dots)$, which we suppress here.

We take the post-trained model $v_\theta$ as the frozen teacher and discretize
$s\in[0,1]$ into a grid $s_0<\cdots<s_M{=}1$. For a clean target $x$---a visual
latent $z_{t+1}$ for the video expert or an action chunk $a_t$ for the action
expert---and noise $\epsilon$, we form the noisy point
$x^{(s_n)}=(1-s_n)\epsilon+s_n x$ and take one teacher Euler step toward the data
end,
\begin{equation}
\hat{x}^{(s_{n+1})} = x^{(s_n)} + (s_{n+1}-s_n)\,
v_\theta\big(x^{(s_n)},\, s_n\big).
\end{equation}
The student is then trained to give consistent predictions at the two adjacent
points,
\begin{equation}
\mathcal{L}_{\text{CD}} = \mathbb{E}_{x,\epsilon,n}\!\left[\,
d\big(f_\xi(x^{(s_n)}, s_n),\;
f_{\xi^-}(\hat{x}^{(s_{n+1})}, s_{n+1})\big)\right],
\label{eq:cd}
\end{equation}
where $\xi^-=\operatorname{stopgrad}(\xi)$ is an exponential-moving-average
target network and $d(\cdot,\cdot)$ is a distance metric (we use squared
$\ell_2$). We distill both experts, reducing the video and action samplers from
$5$ and $10$ steps to $2$ steps each, so the deployed policy produces every chunk
in two function evaluations.

\subsubsection{Inference Acceleration}
\label{ssec:acceleration}

We organize inference acceleration into three levels: model-level compilation, sequence-level attention optimization, and system-level runtime amortization. These levels correspond to different parts of the inference stack. At the model-execution level, we reduce the cost of each DiT forward pass. At the autoregressive sequence level, we optimize how historical states are stored, updated, and attended to during long-horizon rollout. At the runtime system level, we eliminate repeated host-side preparation, redundant memory allocation, and unnecessary synchronization around model invocation.

\paragraph{Model Level: Low-Precision Compiled Execution.}
The dominant computation in each inference step is the DiT forward pass for the video and action branches. We deploy the DiT through an ONNX-based TensorRT build pipeline. TensorRT converts the PyTorch computation graph into an optimized execution plan, performs hardware-specific kernel selection, and fuses compatible operations when supported by the backend. To satisfy the static interface required by TensorRT, runtime states such as the KV cache are represented as explicit engine inputs and outputs, rather than implicit Python-side module states. This interface exposes the cache capacity during engine construction and enables fixed execution profiles. FP8 quantization nodes are inserted with NVIDIA Model Optimizer. In practice, the head and tail layers remain in BF16, while the linear layers inside transformer blocks are executed in FP8.

\paragraph{Sequence Level: Long-Horizon Attention Optimization.}
Although compiled execution accelerates individual DiT calls, autoregressive rollout still faces increasing attention cost as the interaction horizon grows. We therefore optimize the cache layout and attention operator jointly. At runtime, we use paged/ragged KV-cache management to avoid repeated padding and repacking of historical tokens. The buffer capacity is determined by the sliding-window length. Cache updates are performed in place within preallocated buffers, while valid lengths, cache positions, and page metadata are tracked explicitly. To directly consume this compact representation, we replace the generic TensorRT attention subgraph with a FlashInfer-based attention plugin, which executes attention over the paged/ragged cache without materializing padded dense tensors.

\paragraph{System Level: Runtime Overhead Reduction.}
Beyond DiT computation and KV-cache management, the asynchronous inference loop contains repeated host-side preparation. Although each operation is individually small, the cumulative cost becomes non-negligible over many rollout chunks and denoising steps. We amortize these costs by caching reusable runtime objects and skipping redundant work whenever the execution state is unchanged. Specifically, we cache accepted TensorRT bindings and shape configurations, reuse runtime tensors and layer views, precompute cross-attention KV states, cache frame-id and scheduler-step metadata, and avoid redundant cache-update paths after bootstrap. We also remove unnecessary CPU-GPU synchronization points from the inference path.

Together, these three levels address complementary bottlenecks in the inference stack: transformer forward computation, long-horizon attention over historical states, and repeated runtime overhead around model execution.

\section{Data Recipe}
\label{sec:data}

\method is trained with a data recipe that follows the multi-task curriculum of
\cref{sec:method}: general text-to-image pretraining, general text-to-video
pretraining, and finally video-action adaptation on embodied data.

\subsection{General image/video data}
For the general text-to-image and text-to-video pretraining stages, we reuse the
image and video corpora curated for LingBot-Video~\cite{lingbotvideo}. These web-scale image and video data
provide broad appearance and dynamics priors before video-action adaptation.

\subsection{Robot data}
Compared with the first-generation LingBot-VA robot corpus~\cite{VA}, which
contains thousands of hours of manipulation data, the robot stage in
\method retains the same broad mixture of public, semi-public, and internally
collected trajectories, while adding thousands of hours of internal robot
demonstrations.
The retained sources include AgiBot~\cite{agibot}, RoboMind~\cite{robomind},
InternData-A1~\cite{interndata}, Open X-Embodiment/OXE~\cite{oxe} including
DROID~\cite{droid}, UMI-style datasets~\cite{umi,fastumi,mvumi}, and
RoboCOIN~\cite{robocoin}.

All robot data are reprocessed with a unified annotation pipeline. Each long
trajectory is segmented into atomic action clips, and each clip is assigned a
language prompt and a global task instruction using Qwen3.5-397B. This relabeling
step repairs missing or overly generic prompts in earlier versions and makes the
language supervision more consistent across embodiments and datasets.

\subsection{Ego-centric human data}\label{ssec:egodata}
In addition to robot data, we include an internal ego-centric human
manipulation corpus as a scalable source of embodied interaction data. The
corpus contains thousands of hours of first-person manipulation videos
across $65.4$k episodes. The data are balanced across five major
tabletop-oriented environment categories: kitchens, dining tables, vanity
tables, office desks, and tool benches. Across these categories, the corpus
involves more than $600$ operators and over $3.0$k scene-task combinations,
covering daily manipulation skills such as object placement, cleaning,
refilling, assembly, packaging, stationery use, tool use, and cosmetic-item
organization.

Each episode is captured as a single egocentric RGB stream and annotated per
frame with the 6-DoF world-frame root poses of both hands and 22 finger-joint
angles per hand. For ego-centric data, the key additional step is
to align these hand poses with the robot gripper action space. We preserve the two 6-DoF hand root trajectories
and convert the high-dimensional finger-joint trajectories into left/right
scalar parallel-gripper apertures. This yields a compact robot-compatible action
representation while retaining the global motion of both hands. The converted
clips are then incorporated into the same unified annotation and filtering
pipeline as the robot corpus before being used for the human--robot
co-training of \cref{ssec:egocentric}.

\subsection{In-context learning data}\label{ssec:incontextdata}

\begin{figure*}[h]
  \centering
  \includegraphics[width=\linewidth]{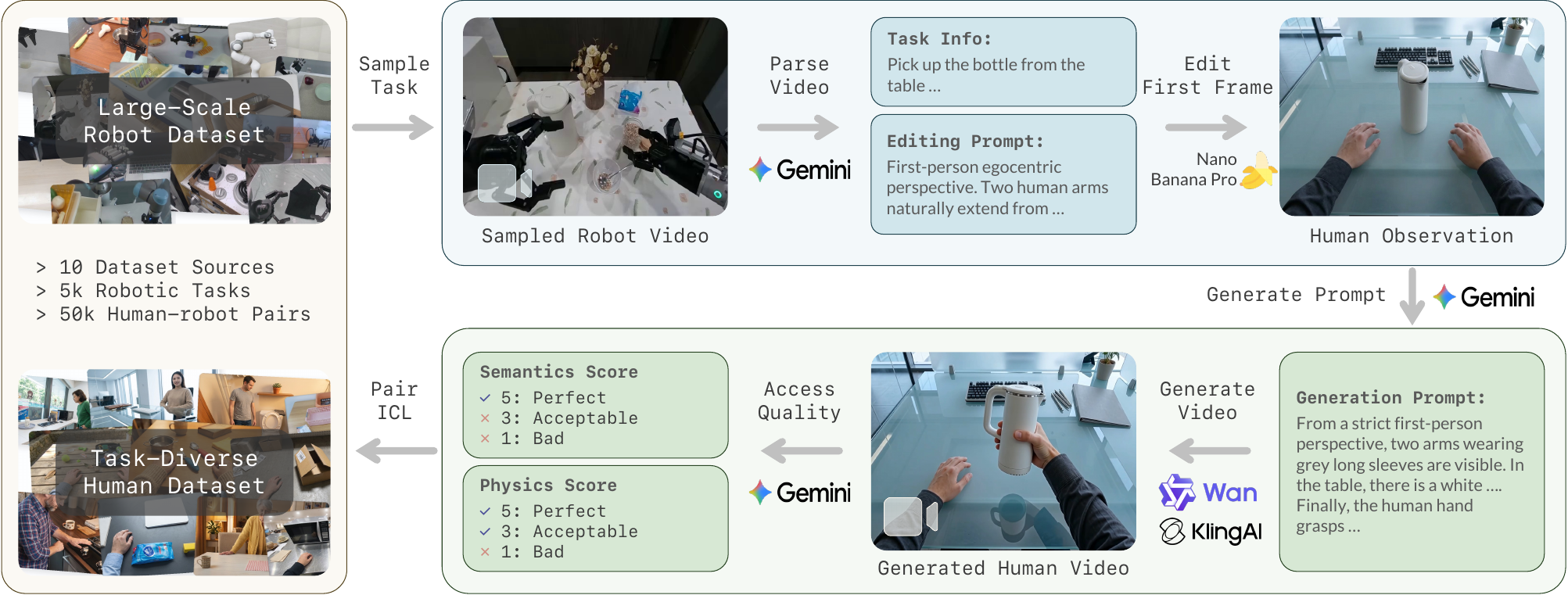}
  \caption{\textbf{In-context human-robot data curation pipeline.} Robot videos
  are sampled by task semantics, converted into human demonstration prompts with
  a VLM, synthesized as human videos, and filtered before pairing with the
  original robot trajectories as ICL samples.}
  \label{fig:icl_data_pipeline}
\end{figure*}

To support human video in-context learning in Section~\ref{ssec:incontext}, we utilize the data pipeline in Zero-WAM~\cite{zerowam} that converts robot videos into human videos that have consistent task semantics. To ensure robust generalization of in-context learning across few-shot or unseen tasks, we sample videos from large-scale robotic video datasets guided by a task taxonomy, thereby constructing a large-scale in-context human-robot dataset featuring rich and diverse task semantics. For each sampled robot video, we first employ a VLM (e.g., Gemini-3.1-Pro~\cite{gemini31pro}) for task analysis and to generate an image editing prompt that transforms the robot's first frame into an initial observation of human manipulation. This editing process imposes no constraints on viewpoint, scene, or object instances, requiring only the preservation of the original robot task semantics. Subsequently, given the edited initial observation image, we leverage the VLM again to generate a human video generation prompt for the corresponding manipulation task, which drives video generation models (e.g., WAN-2.6~\cite{wan26} or Kling-V3~\cite{klingv3}) to synthesize human manipulation videos. Finally, we utilize the VLM to evaluate the generated videos by assigning scores for task semantic preservation and physical plausibility. Qualified human videos are then paired with their original robot counterparts to form individual ICL samples. The resulting ICL dataset encompasses data from over 10 robot video pre-training datasets (e.g., AgiBot~\cite{agibot}, OXE~\cite{oxe}), comprising more than 5,000 tasks and over 50,000 samples.

\section{Experiments}
\label{sec:exp}

\subsection{Implementation \& Training Details}

\subsubsection{Network Architecture}

\paragraph{Backbone.} Our model is a diffusion transformer (DiT) trained from scratch, operating on the latent space of our semantic visual-action tokenizer (96 latent channels). Video latents are patchified with a $1\times2\times2$ ($T\times H\times W$) linear patch embedding into a stream of visual tokens. The video backbone consists of 30 transformer blocks with video hidden dimension 2048; joint self-attention is performed in a 3072-dim attention space (24 attention heads $\times$ 128 head dim). Its feed-forward pathway is implemented as a sparse MoE routed layer with 128 routed SwiGLU experts, top-8 routing, one shared expert, and per-expert intermediate dimension 512, while the action stream uses a dense FFN as described below.

\paragraph{MoT architecture.} Action tokens are processed by a parallel, narrower expert stream inside every block. Raw actions have a unified 30-dim action space (quantile-normalized per dataset; missing dimensions zero-padded and masked) and are embedded by a single bias-free linear layer into an action hidden dimension of 768 (a quarter of the shared attention width). The action FFN dimension is scaled proportionally to 3072. Video and action tokens interact through joint self-attention in the shared 3072-dim attention space: video queries/keys/values are projected $2048\rightarrow3072$, action queries/keys/values are lifted $768\rightarrow3072$, and the attention outputs are projected back to their respective stream widths. In cross-attention the action stream has its own query and output projections but shares the text key/value projections with the video stream. All remaining components are modality-specific: the sparse MoE layer for video, the dense FFN for actions, LayerNorms, AdaLN tables, the timestep/text condition embedder (a 768-dim replica), and the output heads (LayerNorm + linear to $96\cdot4$ latent channels per patch for video; LayerNorm + linear $768\rightarrow30$ for actions).

\paragraph{Causal chunked generation.} The model generates video autoregressively at the chunk level: frames are grouped into temporal chunks, and self-attention is block-causal with a sliding window over past chunks. During training, the chunk size is resampled uniformly from 1--4 latent frames and the attention window from 1--64 chunks at every step, so a single model supports variable chunk sizes and history lengths at inference (we use chunk size 2 and a 64-chunk window for evaluation). To reduce exposure bias during autoregressive rollout, we apply diffusion-forcing-style context noise augmentation: with probability 0.5 the clean history frames are replaced by noised versions at a randomly sampled (small) noise level.

\paragraph{Multi-Chunk Prediction (MCP).} As an auxiliary training objective, lightweight MCP heads predict 1--3 chunks beyond the next chunk. Hidden states are collected from backbone layers $\{3, 11, 19, 29\}$, fused by a two-layer SiLU MLP ($4\times2048\rightarrow2048\rightarrow2048$), and fed to 3 depth groups of 3 extra transformer blocks each (identical configuration to backbone blocks, including the action expert); each depth $d$ additionally receives the noisy latents of the $(d{+}1)$-th future chunk through a per-depth linear projection (concat $\rightarrow$ 2048). The three depths are supervised with loss weights 0.5 / 0.2 / 0.1. MCP heads are used only for training-time supervision; standard inference runs the backbone alone.

\paragraph{Conditioning.} Text prompts are encoded by a dense UMT text encoder into 4096-dim token embeddings (up to 512 tokens), injected via cross-attention after a two-layer MLP projection. Diffusion timesteps enter through sinusoidal embeddings (256-dim) followed by an MLP that produces the AdaLN modulation signals; the video and action streams carry independent timestep conditioning, allowing decoupled noise levels for the two modalities.

\paragraph{Parameter counts (approx.).}
Video backbone ${\approx}13.0$B total / ${\approx}1.9$B active;
action expert ${\approx}0.6$B; MCP heads ${\approx}1.7$B;
total ${\approx}15.3$B trained parameters (of which ${\approx}2.5$B are active
per token at inference).

\subsubsection{Training Recipe}

\paragraph{Objective.} All stages are trained with a rectified-flow (flow-matching) objective: the network predicts the velocity between noise and data, supervised by MSE with the scheduler's timestep-dependent loss reweighting. Timesteps are sampled uniformly with a modality-specific timestep shift: image 2, web/video data 2, robot video--action data 5, MCP branch 10, and 1 (unshifted) for the action modality, whose timestep is sampled independently of the video timestep. Text embeddings are dropped with probability 0.1 for classifier-free guidance.

\paragraph{Optimization.} We use a hybrid Muon + AdamW optimizer. All 2D weight matrices inside transformer blocks (attention projections, dense action-FFN projections, and routed MoE expert projections) are updated with Muon (with lr $2\times10^{-3}$, momentum 0.95, weight decay 0.1); all remaining parameters---embedders, normalization/AdaLN parameters, biases, and output heads---with AdamW (lr $1\times10^{-4}$, $\beta=(0.9, 0.95)$, $\epsilon=10^{-8}$, weight decay 0.01). The learning rate follows a linear warm-up over 2000 steps and is held constant thereafter. Gradients are clipped to a global norm of 1.0.

\paragraph{Systems.} Training is implemented with FSDP full-parameter sharding (no tensor/pipeline/context parallelism), bf16 mixed precision with FP32 gradient reduction, full activation checkpointing, and an elastic multi-node launcher. Each rank processes one packed sequence per step (local batch size 1). We hold out 1\% of the data for validation.

\subsection{Real-world Deployment}

\subsubsection{Real-world Evaluation}
We evaluate real-world manipulation on an in-house robot benchmark covering a
diverse set of everyday manipulation tasks. We report results on four tasks:
i) \emph{Fruit Sorting}, where the robot places an apple, a pear, and an orange
into a basket; ii) \emph{Pen Collection}, where the robot sequentially places
three pens from the table into a cup; iii) \emph{Drawer Tidying}, where the
robot opens a drawer and stores objects from the tabletop inside it; and
iv) \emph{Plate Handover}, where the robot pushes a plate to the center, picks up a
paper pad, places it on the plate, and then places the target object on the
pad. For each task, we collect 20 teleoperated demonstrations and train a
single generalist policy with multi-task supervised fine-tuning, rather than
training one model per task. At evaluation time, the same checkpoint is
deployed across all tasks, and we report the per-task success rate as well as
the average task progress score over repeated rollouts. 

\begin{figure*}[t]
  \centering
  \vbox{%
    \hbox{\includegraphics[width=\linewidth]{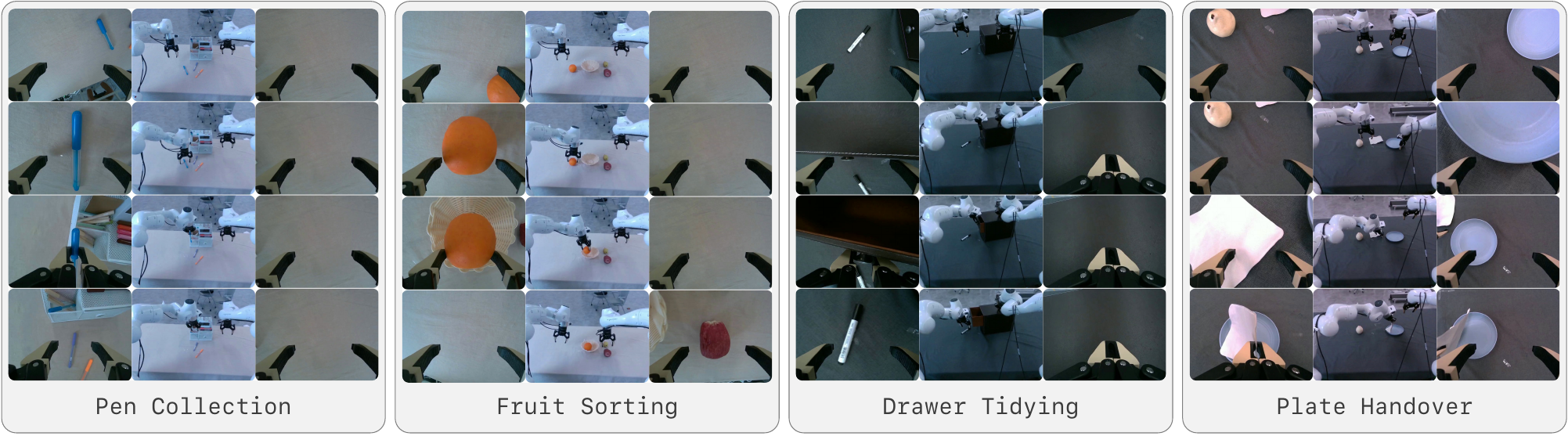}}%
    \nointerlineskip
    \vspace{2pt}
    \hbox{\includegraphics[width=\linewidth]{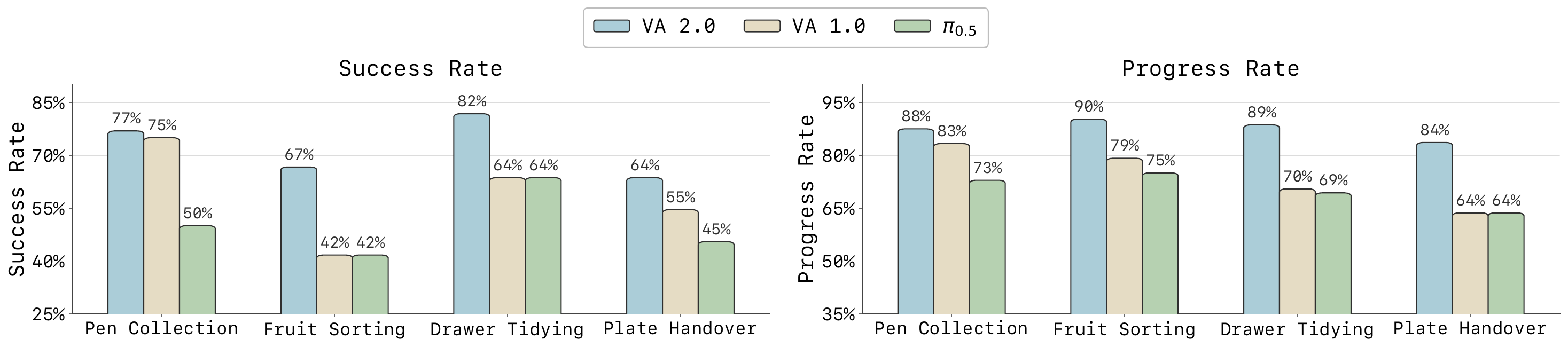}}%
  }
  \caption{\textbf{Real-world deployment results.} Success rate and task progress rate
  across real-world tasks, comparing \method against representative baselines.}
  \label{fig:realworld}
\end{figure*}

We compare against representative baselines from both vision-language-action
(VLA) and video-action (VA) model families. For VLA models, we use
$\pi_{0.5}$~\cite{pi2025pi05} as the main baseline. For VA models, we compare
against LingBot-VA~\cite{VA}, which also models future visual dynamics for
action generation. As shown in \cref{fig:realworld}, \method achieves the
strongest overall real-world performance, improving both success rate and task
progress over the VLA baseline and the prior VA baseline. 
Compared with $\pi_{0.5}$, the improvement is especially clear on continuous
manipulation tasks that require longer-horizon visual grounding. By explicitly
predicting future visual states, \method can maintain task-level context across
multiple interaction steps, while its video grounding also improves the
precision of local manipulation. Compared with LingBot-VA, the gains come from
causal video-action pretraining, which exposes the model to more generalizable
video-action correspondences before downstream policy finetuning. This broader
pretraining is particularly beneficial in the more challenging multi-task
setting, where the policy must generalize across diverse objects, scenes, and
task procedures.

\subsubsection{In-context Learning Demo} 
\begin{figure*}[t]
  \centering
  \includegraphics[width=\linewidth]{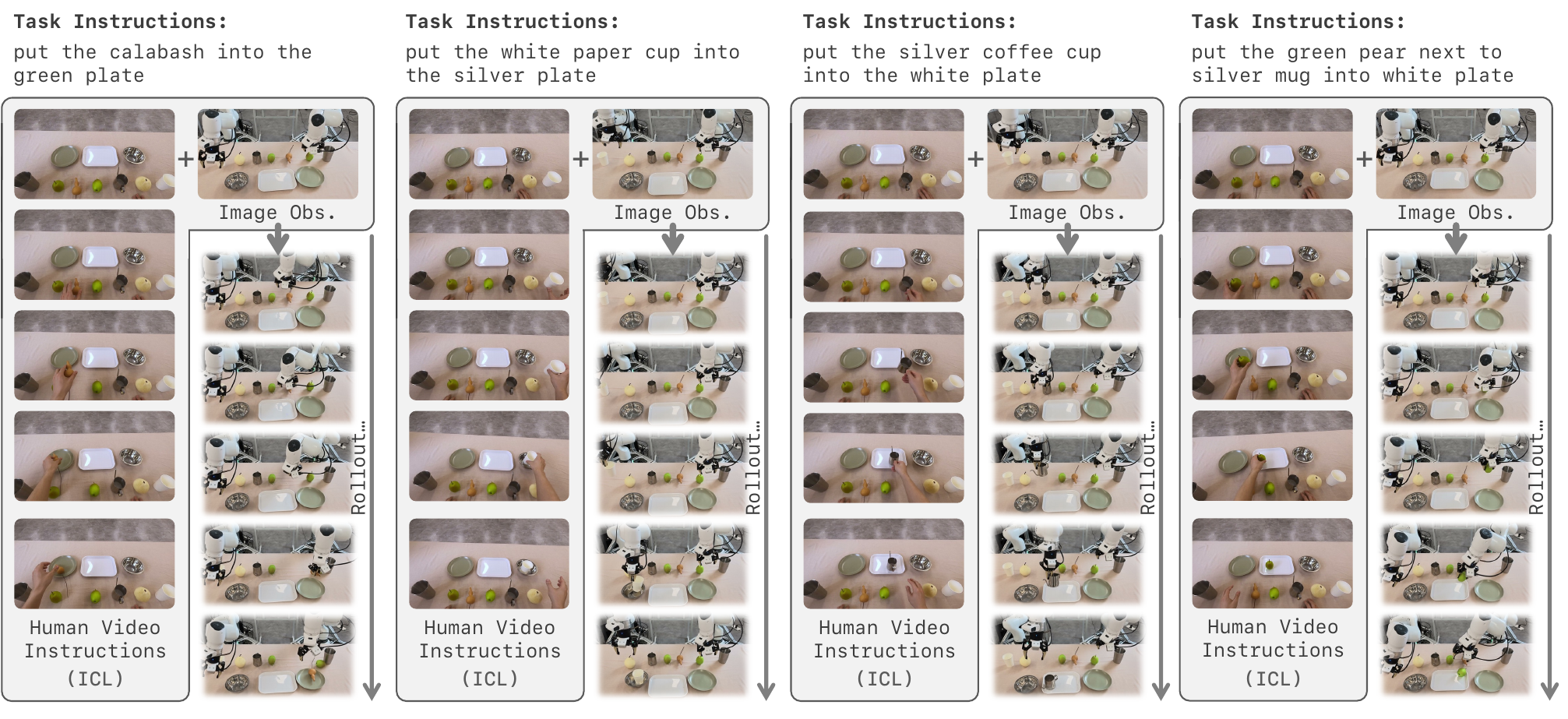}
  \caption{\textbf{In-context learning demo.} We demonstrate ICL rollouts on
  four unseen tasks. For each rollout, given a reference video demonstration as
  the task instruction and the robot observation, \method transfers the same
  action procedure to robot manipulation without parameter updates.}
  \label{fig:icl_demo}
\end{figure*}

To qualitatively examine whether \method can use visual demonstrations as task
context instead of text instructions, we conduct a separate in-context learning
(ICL) study from the real-world evaluation above. As shown in
\cref{fig:icl_demo}, the tabletop scene contains three plates (green, white,
and silver) and seven objects: a silver mug, two green pears, a calabash, a silver coffee cup, a yellow pear, and a white paper cup. For data collection, we only collect
four training tasks: i) \emph{put the white paper cup into the white plate}; ii) \emph{put the silver coffee cup into the silver plate}; iii) \emph{put the green pear in the middle into the green plate}; and
iv) \emph{put the silver mug in the left into the white plate}, with 15  demonstrations per task. We finetune our pretrained model on these four seen tasks.

At test time, we evaluate on new task compositions that are not observed in
the training set. The policy receives a human reference video as the
demonstration context together with the current robot observation, and then
executes the task without updating model parameters. \Cref{fig:icl_demo}
visualizes examples of four unseen tasks: i) \emph{put the calabash into the green plate}, ii) \emph{put the white paper cup into the silver plate}, iii) \emph{put the silver coffee cup into the white plate}, and iv) \emph{put the green pear next to the silver mug into the white plate}, all of which are absent from
the training tasks. This setting evaluates whether the video-action backbone
can extract procedural cues from the context video and transfer them to new task
compositions, object arrangements, and object instances.

\subsection{Simulation}
We evaluate \method on RoboTwin~\cite{robotwin}, a bimanual manipulation
benchmark with clean and domain-randomized settings, and report task success
rate (\%) against representative baselines. Following LingBot-VA~\cite{VA}, we
adopt a multi-task training setup where all models are trained on 2,500
demonstrations collected in clean scenes (50 per task) plus 25,000
demonstrations from heavily randomized scenes (500 per task).

\begin{table}[t]
  \centering
  \caption{\textbf{RoboTwin 2.0.} Bimanual success rate (\%) under clean and
  domain-randomized settings. Clean/Randomized correspond to the Easy/Hard
  settings reported in LingBot-VA~\cite{VA}.}
  \label{tab:robotwin}
  \begin{tabular}{lccc}
    \toprule
    Method & Clean & Randomized & Avg.\ \\
    \midrule
    X-VLA~\cite{xvla}             & 72.9 & 72.8 & 72.9 \\
    $\pi_{0.5}$~\cite{pi2025pi05} & 82.7 & 76.8 & 79.8 \\
    Motus~\cite{motus}            & 88.7 & 87.0 & 87.9 \\
    LingBot-VA~\cite{VA}          & 92.9 & 91.6 & 92.2 \\
    \midrule
    \method (Ours)                & \textbf{93.8} & \textbf{93.4} & \textbf{93.6} \\
    \bottomrule
  \end{tabular}
\end{table}

\paragraph{Results.}
As shown in \cref{tab:robotwin}, \method achieves the best performance on both
clean and domain-randomized RoboTwin settings. Compared with the VLA baseline
$\pi_{0.5}$, \method improves the average success rate by 14.0 percentage
points, indicating that native video-action pretraining provides stronger
physical grounding for bimanual manipulation. Among prior VA methods, LingBot-VA
is the strongest baseline with a 92.2\% average success rate, while \method
further improves the average result to 93.6\%. The gap between clean and
randomized evaluation is also small for \method (0.6 percentage points),
suggesting that the model preserves robust control under visual and physical
domain variation.

\subsection{Ablation}

\begin{table}[t]
  \centering
  \caption{\textbf{Tokenizer ablation on RoboTwin 2.0.} Success rate (\%) of
  a 1.3B video-action model with different visual tokenizers on Easy and Hard
  splits over 50 tasks.}
  \label{tab:vae_ablation}
  \begin{tabular}{lcccc}
    \toprule
    & \multicolumn{2}{c}{WAN2.2 VAE} & \multicolumn{2}{c}{Our tokenizer} \\
    \cmidrule(lr){2-3}\cmidrule(lr){4-5}
    Metric & Easy & Hard & Easy & Hard \\
    \midrule
    Average$_{\mathrm{hor}=1}$ & 81.1 & 78.4 & \textbf{86.2} & \textbf{83.1} \\
    Average$_{\mathrm{hor}=2}$ & 75.5 & 73.9 & \textbf{85.7} & \textbf{84.0} \\
    Average$_{\mathrm{hor}=3}$ & 67.2 & 68.0 & \textbf{92.0} & \textbf{85.4} \\
    \midrule
    Average$_{50\ \mathrm{Tasks}}$ & 78.0 & 76.0 & \textbf{86.6} & \textbf{83.1} \\
    \bottomrule
  \end{tabular}
\end{table}

\paragraph{Semantic Visual-Action Tokenizer.} To isolate the effect of
the tokenizer, we train a 1.3B video-action model from scratch for each variant.
The WAN2.2 VAE baseline and our tokenizer are trained with the same number of
tokens and used to initialize the same downstream architecture. We then
apply identical RoboTwin post-training and evaluate success rate on the official
Easy and Hard splits. The tokenizer ablation in \cref{tab:vae_ablation} shows that replacing the
reconstruction-oriented WAN2.2 VAE with our tokenizer consistently improves
RoboTwin performance across prediction horizons. The gain becomes larger for
longer horizons, suggesting that semantic visual-action tokenization preserves
state information that is more useful for world-action modeling and downstream
control.

\paragraph{Multi-chunk prediction.}
\begin{figure}[t]
  \centering
  \includegraphics[width=\linewidth]{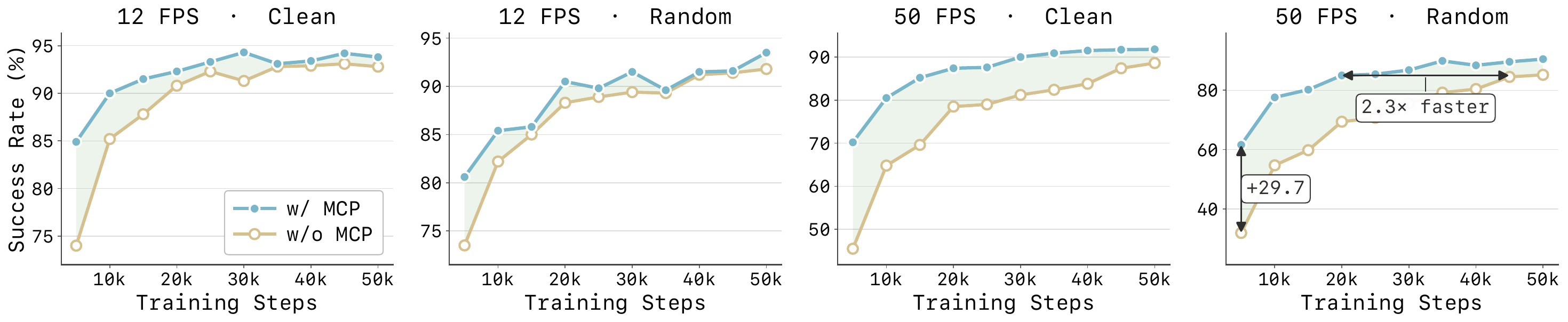}
  \caption{\textbf{Multi-chunk prediction ablation.} Task success rate (\%) on
  RoboTwin across training steps. MCP converges faster and reaches
  higher final accuracy than the baseline at both 12 and 50 fps.}
  \label{fig:ablation_mcp}
\end{figure}

\Cref{fig:ablation_mcp} compares post-training dynamics with and without the
multi-chunk prediction (MCP) objective. We first train a 5B video-action
baseline, then perform RoboTwin post-training under the same data and
optimization setup, with MCP enabled in only one of the two variants. MCP consistently
improves optimization efficiency, converging faster and reaching higher final
task success at both 12 and 50 fps. The advantage is most pronounced at 50 fps:
after 5k training steps, the MCP variant already outperforms the baseline by
29.7 percentage points on the randomized setting, and it matches the baseline's
45k-step accuracy using only 20k steps, corresponding to a 2.3$\times$ training
speedup.

\paragraph{Acceleration.} 
\begin{table}[t]
  \centering
  \caption{\textbf{Inference acceleration.} Inference time and asynchronous
  control frequency under different acceleration techniques.}
  \label{tab:acceleration}
  \begin{tabular}{lcc}
    \toprule
    Acceleration technique & Inference time (ms/chunk) $\downarrow$ & Async Hz $\uparrow$ \\
    \midrule
    BF16 PyTorch async rollout baseline & 927 & 35 \\
    \midrule
    + Consistency distillation & 466 & 69 \\
    + Low-precision compiled execution & 369 & 87 \\
    + Long-horizon attention optimization & 272 & 118 \\
    + Runtime overhead reduction & \textbf{142} & \textbf{225} \\
    \bottomrule
  \end{tabular}
\end{table}

\Cref{tab:acceleration} reports the cumulative effect of our inference acceleration pipeline.
The Async Hz values are computed as $(1000 / t_{\mathrm{chunk}}) \times K$, where $t_{\mathrm{chunk}}$ is the measured inference time in milliseconds and $K$ is the number of low-level control steps in each generated chunk; we set $K=32$ in our experiments.
Starting from a BF16 PyTorch asynchronous rollout baseline, consistency distillation first reduces the number of denoising steps required at each chunk, decreasing the inference time from 927 ms to 466 ms per chunk. We then apply the three inference optimizations described in \Cref{ssec:acceleration}. Low-precision compiled execution accelerates the model-execution layer by replacing eager DiT forward passes with FP8 TensorRT engines, reducing latency to 369 ms per chunk. Long-horizon attention optimization further improves the autoregressive sequence layer by using paged/ragged KV-cache execution and FlashInfer attention plugins, reducing latency to 272 ms per chunk. Finally, runtime overhead reduction amortizes repeated host-side preparation, memory allocation, and synchronization costs, bringing the end-to-end inference time to 142 ms per chunk. Overall, these optimizations improve the asynchronous control frequency from 35 Hz to 225 Hz, yielding a 6.5$\times$ end-to-end speedup.

\subsection{Demonstration}
\label{subsec:demonstration}

\begin{figure*}[t]
  \centering
  \includegraphics[width=\linewidth]{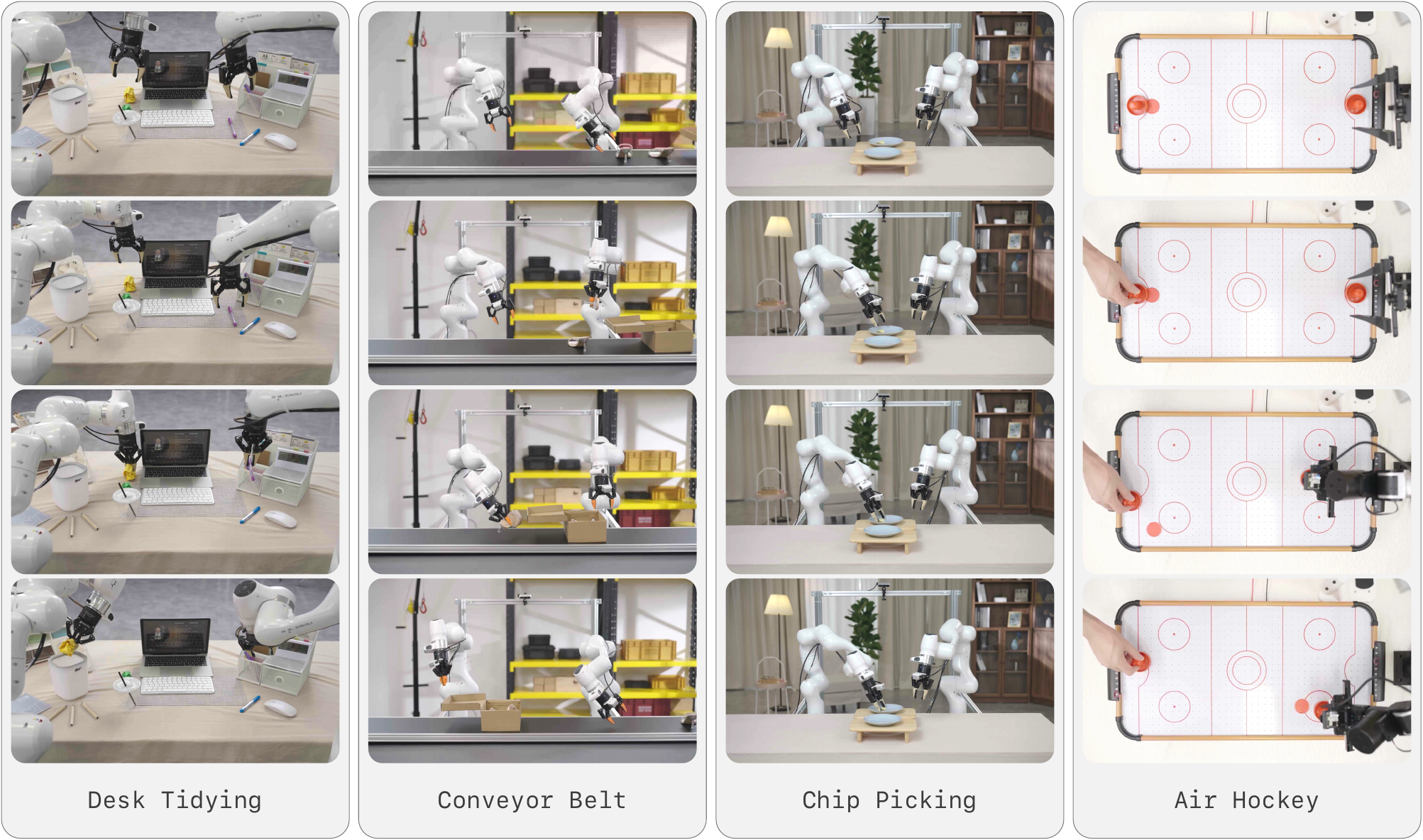}
  \caption{\textbf{Real-world demonstrations.} Qualitative rollouts of \method
  across four representative tasks, covering long-horizon tabletop
  organization, interaction with moving objects, fine-grained grasping, and
  reactive visual control.}
  \label{fig:demo}
\end{figure*}

\Cref{fig:demo} shows four real-world demonstrations that exercise different
aspects of video-action modeling. \textbf{Desk Tidying} emphasizes long-horizon
planning, requiring the policy to parse a cluttered tabletop scene, maintain
object state over time, and sequence a multi-step cleanup behavior.
\textbf{Conveyor Belt} tests temporal grounding and dynamic prediction, where
the model must anticipate moving objects and synchronize each action chunk with
the conveyor motion. \textbf{Chip Picking} stresses fine-grained manipulation
and control without any tactile sensors, requiring visually guided, gentle
grasping of a thin object. \textbf{Air Hockey} evaluates fast inference and
reactive control in a dynamic game setting, where the policy must quickly
predict the puck trajectory and close the loop from recent observations.

\section{Related Work}
\label{sec:related}

\subsection{Generalist robot policies and VLA models}
Vision-language-action (VLA) models directly map visual observations and language
instructions to actions, inheriting broad semantic priors from internet-scale
pretraining~\cite{rt1,rt2,kim2024,roboflamingo,octo,rdt,pi0,pi2025pi05,pi07,xvla,gr3,gr00t,geminirobotics,cogact,hybridvla,dexvla,magma,nora,barreiros2025careful}.
Industrial foundation models such as LingBot-VLA~\cite{lingbotvla},
Qwen-RobotManip~\cite{qwenrobotmanip}, and WALL-OSS
variants~\cite{walloss,walloss05} further push these policies toward real-world
deployment.
A closely related line learns visuomotor policies that regress action sequences
from demonstrations, such as ACT~\cite{act} and Diffusion
Policy~\cite{diffusionpolicy}, while others add visual chain-of-thought or
instruction reasoning~\cite{zhao2025cot,yang2025_1,f1,hi_robot} and inject
temporal context through explicit memory
modules~\cite{li2025cronusvla,sridhar2025memer,shi2025a}. These policies are
reactive: they learn a direct
observation-to-action mapping without modeling how the scene evolves, which
limits long-horizon consistency and sample efficiency. In contrast, \method
grounds control in predicted physical dynamics.

\subsection{Video generation and world-action models for control}
World-action models extend visual world modeling to robot control, using
predicted scene evolution as an intermediate representation for action
generation~\cite{VA,dreamzero,dreamgen,worldvla,threedvla,gr2,gr1,seer,fastwam,ivideogpt,irasim,vidman,adaworld,enerverse,genieenvisioner,nwm,wallwm}.
DreamZero~\cite{dreamzero} adapts a large pretrained video diffusion model for
joint video-action modeling, while LingBot-VA~\cite{VA} formulates training as
causal world modeling that jointly learns frame prediction and policy execution.
A broader body of work couples video prediction with policy learning from
complementary angles, including video generation as policy and video-conditioned
policies~\cite{unipi,gen2act,dreamitate,vpp,avdc} and joint video-action
modeling~\cite{uwm,uva}. Related efforts treat interactive or controllable video
generation as a dynamics model~\cite{genie2,gamegenx,onexwm,lingbotworld,lingbotworldv2} and use generated
rollouts as simulators or imagined plans for policy learning and
evaluation~\cite{kim2026cosmos,team2025evaluating,zhou2024robodreamer,du2023learning,zhou2025act2goal,liu2025a,wang2025a}.
Beyond pixel-space prediction, world models for control also operate in compact
latent spaces~\cite{dreamerv3,daydreamer,tdmpc2,watter2015embed,li2019propagation,shen2024action,lusch2018deep}
or over structured 3D and particle
states~\cite{li2024deformnet,xu2019densephysnet,shi2024robocraft,wang2023dynamic,sulsky1995application,zhang2024adaptigraph,zhang2025particle}.
Training such causal video models also builds on autoregressive
video-generation techniques that close the train--test gap, such as Diffusion
Forcing~\cite{diffusionforcing} and Self Forcing~\cite{selfforcing};
flexible multimodal diffusion frameworks~\cite{unidiffuser} offer complementary
conditioning. Crucially, these methods adapt off-the-shelf,
bidirectional video generators~\cite{sora,wan2024video,veo} through continued
training, whereas \method pretrains a native, causal video-action stack from
scratch.
Direct Video-Action (DVA)~\cite{rhoda_dva} is closely aligned with \method in
advocating native causal video-action pretraining rather than retrofitting a
bidirectional video generator for control. Both approaches treat action
generation as part of a causal visual dynamics model, so the policy learns to
predict how the scene will evolve while producing controls. Our design differs
in two core components. First, we build the video-action stack on a semantic
visual-action tokenizer whose VAE is explicitly aligned for control, tying visual
states and latent actions in a shared representation rather than relying on a
generic reconstruction-oriented video latent. Second, we design a sparse
Mixture-of-Experts architecture for the video stream, increasing dynamics
modeling capacity while keeping the action stream dense and stable for control.
Together with our inference-time acceleration and re-grounded asynchronous
rollout, these choices make native causal pretraining better matched to
long-horizon robot manipulation while enabling faster inference and more
responsive closed-loop control.

\subsection{Representations for control: semantic tokenizers and latent actions}
Another line recovers action-like variables directly from observation sequences,
making unlabeled video usable for control~\cite{vpt}. Genie~\cite{genie} makes video
dynamics controllable through latent actions; LAPO~\cite{lapo} recovers
latent-action policies, world models, and inverse dynamics purely from videos;
LAPA~\cite{lapa} scales discrete latent actions to robot manipulation; and
Moto~\cite{moto} and UniVLA~\cite{univla} shift the abstraction toward motion
tokens. RepWAM~\cite{repwam} and Motus~\cite{motus} pair such latent actions with
representation-oriented tokenizers. \method builds on this direction but places
world states and actions in a single semantic latent space---aligning visual
latents to a foundation model and learning latent actions jointly---rather than
relying on a reconstruction-only VAE with a separately attached action module.

\section{Conclusion}
\label{sec:conclusion}

We presented \method, a video-action foundation model built natively for robot
control rather than adapted from generic video generation. Its design follows
one principle throughout: every component, from representation to inference, is
shaped by the demands of embodied control. A semantic visual-action tokenizer
places world states and latent actions in a single latent space aligned with a
visual foundation model; a causal DiT with a sparse Mixture-of-Experts video
stream is pretrained from scratch on this space, with multi-chunk prediction,
in-context learning, and human--robot co-training enriching the training
signal; and Foresight Reasoning, together with few-step distillation and
system-level acceleration, turns the pretrained model into a real-time
closed-loop controller. Across simulation and real-world evaluations, \method
improves over strong vision-language-action and video-action baselines, adapts
to new tasks from only a few demonstrations, and sustains long-horizon
manipulation through closed-loop re-grounding and hierarchical planning.

Several directions remain open. The planner and the policy are currently
trained separately and coupled through a fixed interface; tighter joint
training may further improve long-horizon consistency. The latent-action space
is learned from passive video, and interactive or reinforcement signals could
sharpen it toward control. Finally, scaling the pretraining corpus and the
sparse backbone further, and broadening embodiment coverage beyond bimanual
manipulation, are natural next steps toward a general-purpose embodied
foundation model.

\section{Contributors}
\label{sec:contributors}

\noindent\textbf{\large Core Contributors}

\vspace{4pt}
\noindent\textbf{Pretraining:} Qihang Zhang, Lin Li, Junke Wang, Jiahao Shao, Gangwei Xu, Luyao Zhang, Jiaming Zhou, Yudong Jin, Shuailei Ma, Jiaqi Liao, Ka Leong Cheng \\
\textbf{Pretraining Data:} Lin Li, Luyao Zhang, Qihang Zhang, Jiaming Zhou, Yudong Jin, Xinyang Wang \\
\textbf{Posttraining:} Luyao Zhang$^{*}$, Shuai Yang$^{*}$, Lin Li, Qihang Zhang, Gangwei Xu, Jiapeng Zhu, Yujie Zhao, Weixuan Tang \\
\textbf{Acceleration:} Shuaiting Li, Chaojian Li \\
\textbf{Deployment \& Demo:} Yiming Luo$^{*}$, Shuai Yang$^{*}$, Ruilin Wang, Yishu Shen, Jiahao Shao, Fangyi Xu, Shuaiting Li, Guanxing Lu, Zifan Shi, Yongkun Wen  \\
\textbf{Project Supervision:} Yujun Shen, Xing Zhu, Nan Xue \\
\textbf{Project Lead:} Yinghao Xu

\vspace{8pt}
\noindent\textbf{\large Contributors}

\vspace{4pt}
\noindent Zelin Gao, Wei Wu, Kecheng Zheng, Ruonan Zhang, Han Zhang, Jingmei Zhao, Yongtao Huang, Haitao Wang, Jingjing Wang, Chen Song

\vspace{6pt}
\noindent{\footnotesize $^{*}$Equal contribution}

{
\small
\bibliographystyle{plain}
\bibliography{ref.bib}
}

\end{document}